\documentclass{article}

\PassOptionsToPackage{numbers, compress}{natbib}

\usepackage[final]{neurips_2022}

\usepackage[utf8]{inputenc} 
\usepackage[T1]{fontenc}    
\usepackage{hyperref}       
\usepackage{url}            
\usepackage{booktabs}       
\usepackage{amsfonts}       
\usepackage{amsmath}
\usepackage{nicefrac}       
\usepackage{microtype}      
\usepackage{xcolor}         
\usepackage{enumitem}
\setlist[itemize]{leftmargin=*}
\setlist[enumerate]{leftmargin=*}

\usepackage{dsfont}
\usepackage{subcaption}
\usepackage{caption}
\usepackage{sidecap}
\usepackage{import}
\usepackage{graphbox}
\usepackage{tikz}
\usepackage{placeins}
\usepackage{verbatim}

\usepackage{etoolbox}
\providetoggle{comments}
\settoggle{comments}{true}

\newcommand{\removetext}[1]{{}}

\newcommand*{\prob}{\mathbb{P}}

\newcommand{\R}{\mathbb{R}}

\newcommand{\indic}{\mathds{1}}

\newtheorem{theorem}{Theorem}[section]

\newtheorem{definition}[theorem]{Definition}

\title{Are Two Heads the Same as One? Identifying Disparate Treatment in Fair Neural Networks}

\author{%
  Michael Lohaus\thanks{Work done during an internship at Amazon Web Services.} \\
  University of Tübingen\\
  Tübingen, Germany\\
  \texttt{michael.lohaus@uni-tuebingen.de} \\
  \And
  Matthäus Kleindessner \\
  Amazon Web Services \\
  Tübingen, Germany \\
  \texttt{matkle@amazon.com} \\
  \AND
  Krishnaram Kenthapadi \\
  Fiddler AI \\
  Palo Alto, USA \\
  \texttt{krishnaram@fiddler.ai} \\
  \And
  Francesco Locatello \\
  Amazon Web Services \\
  Tübingen, Germany \\
  \texttt{locatelf@amazon.com} \\
  \And
  Chris Russell \\
  Amazon Web Services \\
  Tübingen, Germany \\
  \texttt{cmruss@amazon.com} \\
}

\begin{document}

\newpage

\maketitle

\begin{abstract}
We show that deep networks trained to satisfy demographic parity often do so through a form of race or gender awareness, and that the more we force a network to be fair, the more accurately we can recover race or gender from the internal state of the network. Based on this observation, we investigate an alternative fairness approach: we add a second classification head to the network to explicitly predict the protected attribute (such as race or gender) alongside the original task. After training the two-headed network, we enforce demographic parity by merging the two heads, creating a network with the same architecture as the original network. We establish a close relationship between existing approaches and our approach by showing (1) that the decisions of a fair classifier are well-approximated by our approach, and (2) that an unfair and optimally accurate classifier can be recovered from a fair classifier and our second head  predicting the protected attribute. We use our explicit formulation to argue that the existing fairness approaches, just as ours, demonstrate \emph{disparate treatment} and that they are likely to be unlawful in a wide range of scenarios under US~law.
\end{abstract}

\section{Introduction}\label{sec:introduction}

Autonomous systems that make substantive decisions about people must  conform to relevant anti-discrimination legislation. Within the US legal system, two common tests of  anti-discrimination legislation are referred to as disparate treatment and disparate impact \citep{king2020blurred}. Disparate treatment corresponds to the idea that people should not be treated differently because of their membership of a protected group (such as race or gender), while disparate impact refers to the idea that seemingly neutral practices should not cause a substantial adverse impact to a protected group. Importantly, it has been argued \cite{barocas2016big} that there are a large range of scenarios where disparate treatment is unlawful even when performed as a remedy to disparate impact. This is in  departure from the EU where more latitude exists when rectifying indirect discrimination (analogous to disparate impact) \citep{wachter2021fairness}.

Consequentially, disparate treatment doctrine prevents a wide range of actions intended to address sustained inequality \cite{bent2019algorithmic}. Of particular relevance to our work is a 1991 amendment to Title VII \cite{civilrights_title7}. This amendment explicitly prohibits the ``use [of] different cutoff scores for . . . employment related tests on the basis of race, color, religion, sex, or national origin'', even if done for reasons of affirmative action. In this paper, we examine the relationship between existing methods for enforcing demographic parity~(a definition closely related to avoiding disparate impact) that are commonly claimed to not exhibit disparate treatment and methods for demographic parity that alter the cutoff score on the basis of inferred race or gender. 
In doing so, we raise fundamental questions about the legality of  existing approaches for enforcing demographic parity.

\begin{figure}[t]
\centering
\includegraphics[width=.95\textwidth]{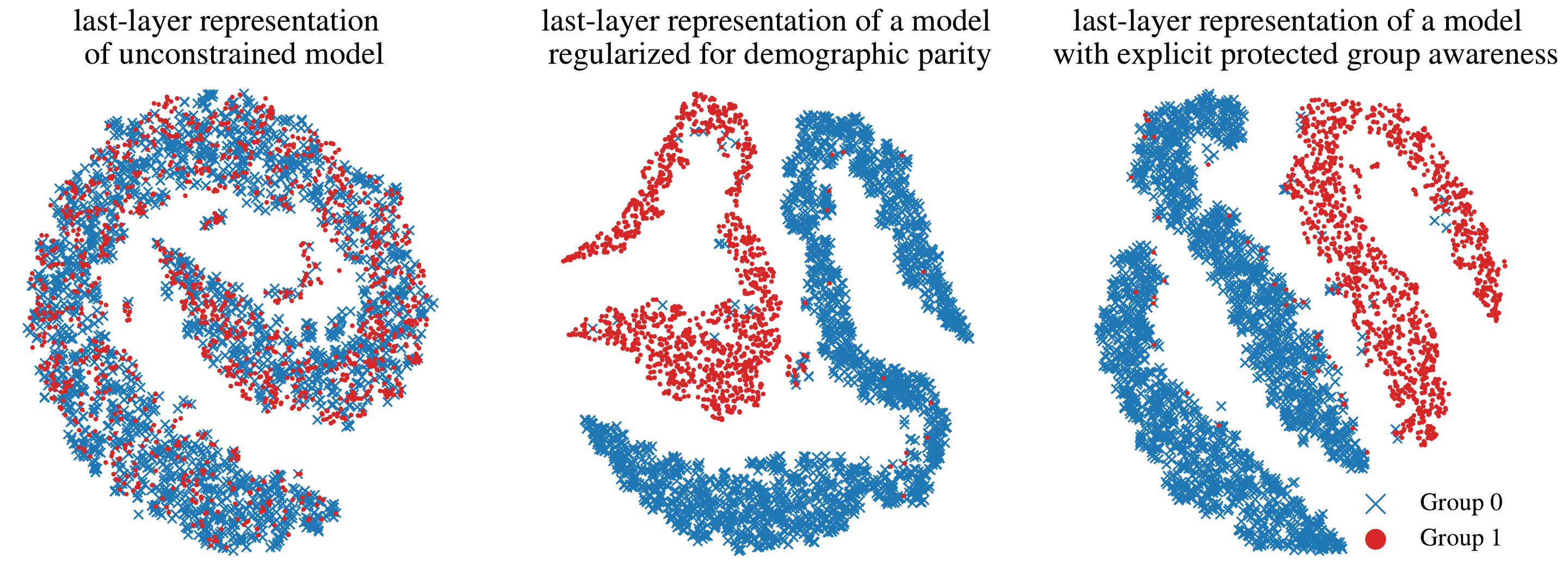}
\caption{{\bf Feature representations of unconstrained (\textit{left}), fairness-regularized (\textit{center}), and group-aware (\textit{right}) ResNet50 models.} The plots show tSNE~\cite{maaten2008tsne} embeddings of 
last-layer representations of the CelebA dataset \cite{liu2015faceattributes}; 
red points correspond to  male individuals, blue points to female ones. Classifiers are trained to identify if people are smiling. The  use of fairness constraints during training causes a mixed manifold  (\textit{left}) to separate into largely disjoint sets (\textit{center}). Similar behavior is observed when training a two-headed model (\textit{right}) that predicts  gender and~`smiling'.
} \label{fig:last_layer_tSNE}
\end{figure}

In particular, we examine the behavior of deep neural networks trained to satisfy 
demographic parity either with a fairness regularizer \citep{wick2019} or by  preprocessing~\cite{Kamiran2012}. Such models could learn an internal representation that  either \emph{(i)}~is independent of the protected attribute (as methods from fair representation learning aim to do---see Appendix~\ref{sec:related_work} for related work), or \emph{(ii)}~that  distinguishes between groups so as to tune ``separate'' classifiers for each group in a way that results in the demographic parity of the network.
In the context of US law, it is vital to understand which case occurs in practice. If the learned algorithms treat people differently on the basis of their race or gender, this may correspond to \emph{disparate treatment}. We find that networks trained to satisfy demographic parity via a fairness regularizer or preprocessing fall into the second case. Figure~\ref{fig:last_layer_tSNE} provides an illustration of this finding. Moreover, we find that the more strongly demographic parity is enforced, the more predictive the internal representation is of the protected attribute.

Building on the observation that the protected attribute is implicitly learned, 
we train a neural network with a second classification head that explicitly predicts the protected attribute. After training, we use the second head response to directly reduce demographic disparity. 
On certain tasks we show that our approach is tightly related to existing regularizer and preprocessing approaches and conclude that these approaches, just as ours,  demonstrate disparate treatment.

To summarize, we make the following contributions:

\begin{enumerate}
    \item  The internal state of a fair network trained with a fairness regularizer or preprocessing  is strongly predictive of the protected attribute. As the fairness is more strongly enforced, the accuracy of predicting the protected attribute increases (Section~\ref{sec:implicit_awareness}).
    \item Merging a second classification head with the network is an effective approach for creating demographic parity-fair classifiers. We show that the performance of our approach closely tracks the optimal accuracy-fairness trade-offs of \citet{lipton2018}; however, in contrast to their method, ours does not require explicit access to the protected attribute (Section~\ref{sec:explicit_approach}).
    \item  As our principal contribution, we show 
    cases where the decisions of a fair model (via  regularizer or preprocessing) are well-approximated by our approach. Similarly, the decisions of an unconstrained classifier can be approximated by a weighted sum of the fair classifier and our second head     (Section~\ref{sec:model_equivalence}).
    \item Using the close relationship between a demographic parity-fair classifier and our explicit approach
    we are able to identify individuals who are systematically disadvantaged by the fair classifier, and who would receive a positive decision if their apparent race or gender were different. This allows us to conclude that the fair method     exhibits disparate treatment (Section~\ref{sec:disparate_treatment}).
\end{enumerate}

\section{Preliminaries}\label{sec:preliminaries}

A binary classifier is a function $h: \mathcal{X}\rightarrow \mathcal{Y}$ that, given a datapoint~$x$ from feature space $\mathcal{X}$, aims to accurately predict the datapoint's assigned label~$y\in \mathcal{Y} = \{0,1\}$.  We  consider thresholded classifiers of the form~$h(x) = \indic(f(x) > 0)$, where $f$ is a continuous function $f:\mathcal{X} \rightarrow \R$. We want $h$ to 
be as accurate as possible, while at the same time being fair with respect to a protected attribute~$s\in \mathcal{S} = \{0,1\}$ 
according to the fairness notion of demographic parity.

\begin{definition}[Demographic Parity]\cite{Kamiran2012,fta2012}
A classifier~$h: \mathcal{X}\rightarrow \{0,1\}$ satisfies \emph{demographic parity} under distribution~$\prob$ if its prediction~$h(X)$ is independent of the protected attribute~$S$,~i.e., 
\begin{align*}
   \prob( h(X) = 1| S=0) = \prob(h(X) = 1| S=1).  
\end{align*}
We measure the violation of demographic parity by the \emph{demographic disparity (DDP)}, given by 
\begin{align}\label{definition_DDP}
    \mathrm{DDP} = \mathrm{DDP}(h; \prob) :=  \prob( h(X) = 1| S=1) - \prob( h(X) = 1| S=0).
\end{align}
\end{definition}

In this paper, we examine two popular fairness methods for training deep neural networks to satisfy demographic parity:

\textbf{Regularized Fair Classification} One common approach for learning a fair classifier is to add a  regularizer to the standard training objective \citep[e.g.,][]{beutel2019fairness,wick2019,padala2020fnnc,lohaus2020, kleindessner2021, risser2021tackling, bendekgey2021scalable}. Such regularizers are used to enforce numerous fairness definitions and typically impose a continuous relaxation of a discrete fairness measure such as the DDP \eqref{definition_DDP}. The regularizer used in~\citep{wick2019} is a sigmoid-based relaxation of 
the squared value of \eqref{definition_DDP}, evaluated on a dataset 
$(x_i,y_i,s_i)_{i=1}^n$ with i.i.d. samples from~$\prob$:
\begin{align}\label{def_fairness_regularizer}
      \widehat{\mathcal{R}}_{\mathrm{DP}}(f) := \Big(\frac{1}{|\{i: s_i=1\}|} \sum_{i \in [n]: s_i = 1}\limits \sigma(f(x_i))- \frac{1}{|\{i: s_i=0\}|} \sum_{i \in [n]: s_i = 0}\limits \sigma(f(x_i))\Big)^2.
\end{align}

The fairness regularizer is added to the overall loss and trades-off fairness versus accuracy via a multiplicative hyperparameter~$\lambda$, where higher values of lambda encourage an increase in fairness at the cost of accuracy. In the paper, we report experimental results for $\widehat{\mathcal{R}}_{\mathrm{DP}}(f)$. Appendix \ref{suppsec:extended_results}  presents results for a related regularizer.
Note that we follow~\citep{bendekgey2021scalable,wick2019} in applying the regularizer~\eqref{def_fairness_regularizer} to the sigmoid output of the networks. This differs from approaches such as \citep{zafar2017fairness,donini2018} that enforce relaxed fairness constraints on margin distances and have recently been criticized for their ineffectiveness~\citep{lohaus2020}. 

\textbf{Preprocessing: Massaging the Dataset} We also examine the preprocessing method of~\citep{Kamiran2012}, which alters target labels prior to training. \emph{Massaging} `promotes' negative points from the disadvantaged group to the positive label if they have a (comparatively) high positive class probability {as calculated by an unconstrained classifier} and `demotes' positive points from the advantaged group to the negative label if they have a low positive class probability. The number of changed labels is controlled by a 
parameter~$\lambda\in[0,M]$, where $\lambda=M$ results in the same fraction of positive points for both~groups. 

\textbf{Fair Representation Learning} methods learn data representations such that any ML model trained on top of such a representation would be fair \cite{zemel2013,madras2018,Beutel2017DataDA,adel2019,zhao2019inherent,edwards2016,louizos2016,feng2019,moyer2018invariant,xie2017controllable,jia2018right,raff2018gradientreversal,alvi2019a}. These techniques come in various flavors, and are considered preprocessing or in-processing methods. When used to train demographic parity-fair classifiers, these methods try to learn representations that  contain no information of the protected attribute. Consequently, these techniques should not show the phenomenon of attribute awareness we found for networks regularized to satisfy demographic parity or trained on massaged datasets, at least not when examining the final representation layer. However, it is an interesting question for future work whether methods for fair representation learning suffer from attribute awareness in earlier~network~layers. Like us, adversarial approaches like \cite{madras2018, Beutel2017DataDA, adel2019,edwards2016, edwards2016} train a second head, but in contrast to our approach they are trained adversarially in a minmax formulation, while the objectives of our two heads are jointly minimized in a multitask setting.
 
In general these fair representation learning methods have noticeably different behavior to preprocessing or regularized fairness methods. Because they try to solve a harder problem of learning a representation that guarantees the fairness of any downstream classifier, rather than making a single classifier fair, they typically show worse accuracy trade-offs than other methods. As such, while we did not did not attempt the analysis set out in Section 5 on these methods, we do not expect our fairness results to hold for them.

\vspace{3mm}
To set the ground for our paper, we now discuss the closely related work of \citet{lipton2018}. We discuss further related work in Appendix~\ref{sec:legal} and~\ref{sec:related_work}. 

\textbf{Implicit Disparate Treatment} \citet{lipton2018} examine the popular claim that machine learning models that do not use protected information at test time do not exhibit disparate treatment~\citep{goh2016,zafar2017www,zafar2017fairness,donini2018,wu2019,padala2020fnnc, harned2019stretching}. \citeauthor{lipton2018} recommend caution and observe that if the protected attribute~$s$ is a deterministic function $s=g(x)$ of the non-protected features~$x$, any sufficiently powerful ML model can learn a function~$f(x,s)=\tilde{f}(x)$ with $\tilde{f}(x)=f(x,g(x))$. They argue that even though the protected attribute is not provided at test time, such a model would constitute a case of disparate treatment since it makes decisions based on the \emph{implicitly reconstructed} protected attribute. However, beyond a synthetic experiment in which a classifier discriminates based on hair length (as a proxy for gender), they do not study whether---or \emph{how}---implicit disparate treatment happens in practice. We provide strong evidence that deep neural networks that are enforced to satisfy  demographic parity by means of a regularizer or preprocessing suffer from disparate treatment, even when not explicitly using the protected attribute at test time---and that they do so by separating last-layer representations based on protected attributes. 

A second contribution of \citet{lipton2018} is to 
prove that the most accurate classifier among all classifiers satisfying a bound on the demographic disparity utilizes group-specific thresholds (a similar result has also been shown by \citet{Menon2018}). They then propose a postprocessing method that greedily chooses per-group decision thresholds for a  classifier that has been learned without fairness constraints; this requires access to protected attributes at test time. In~Section~\ref{sec:explicit_approach} we compare our proposed two-head approach for demographic parity-fairness to their approach.

\section{Protected Attribute Awareness in Fair Networks}\label{sec:implicit_awareness}

\begin{figure}
\centering
\includegraphics[height=7.25cm]
{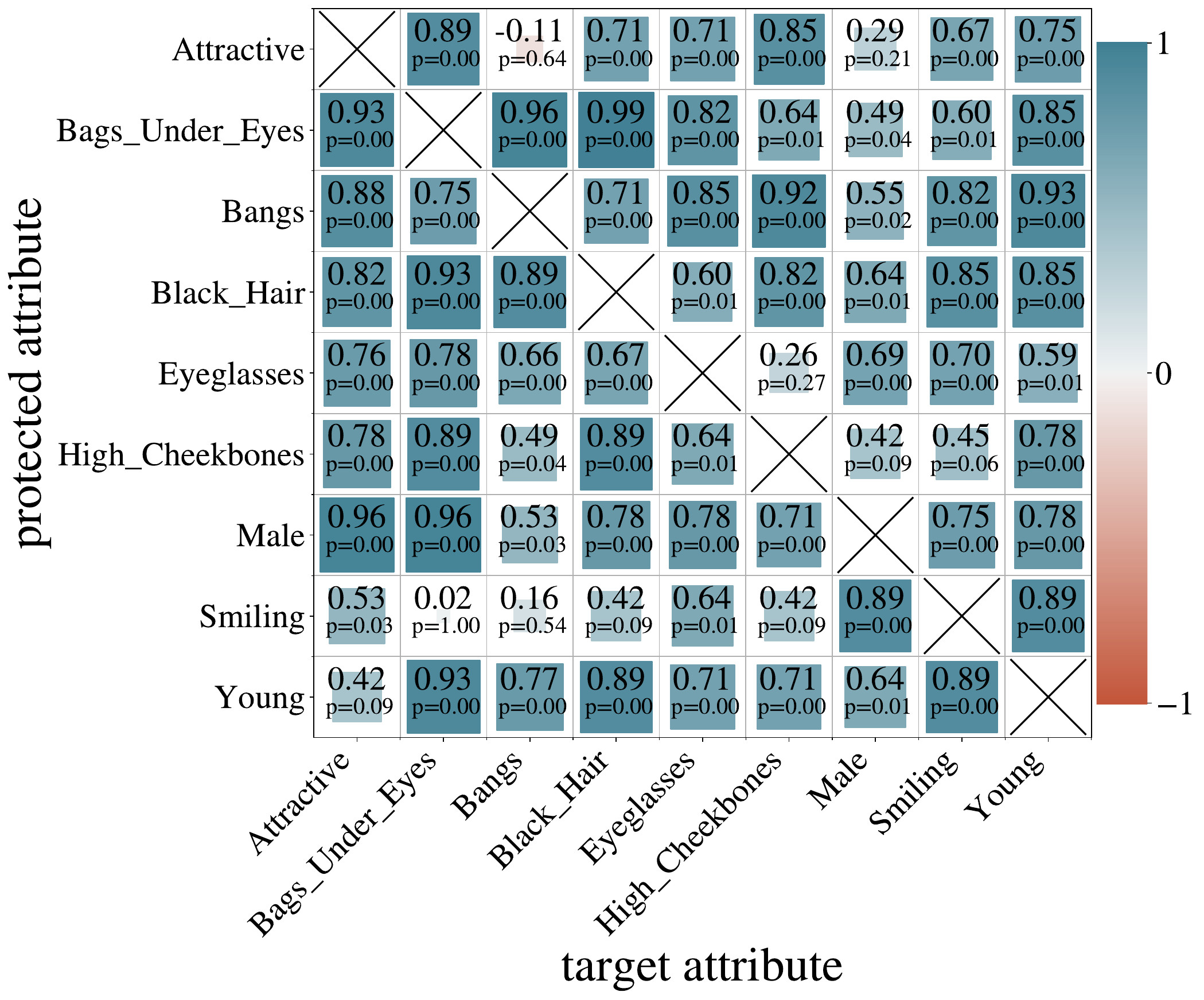}
\hfill
\includegraphics[height=7.25cm]
{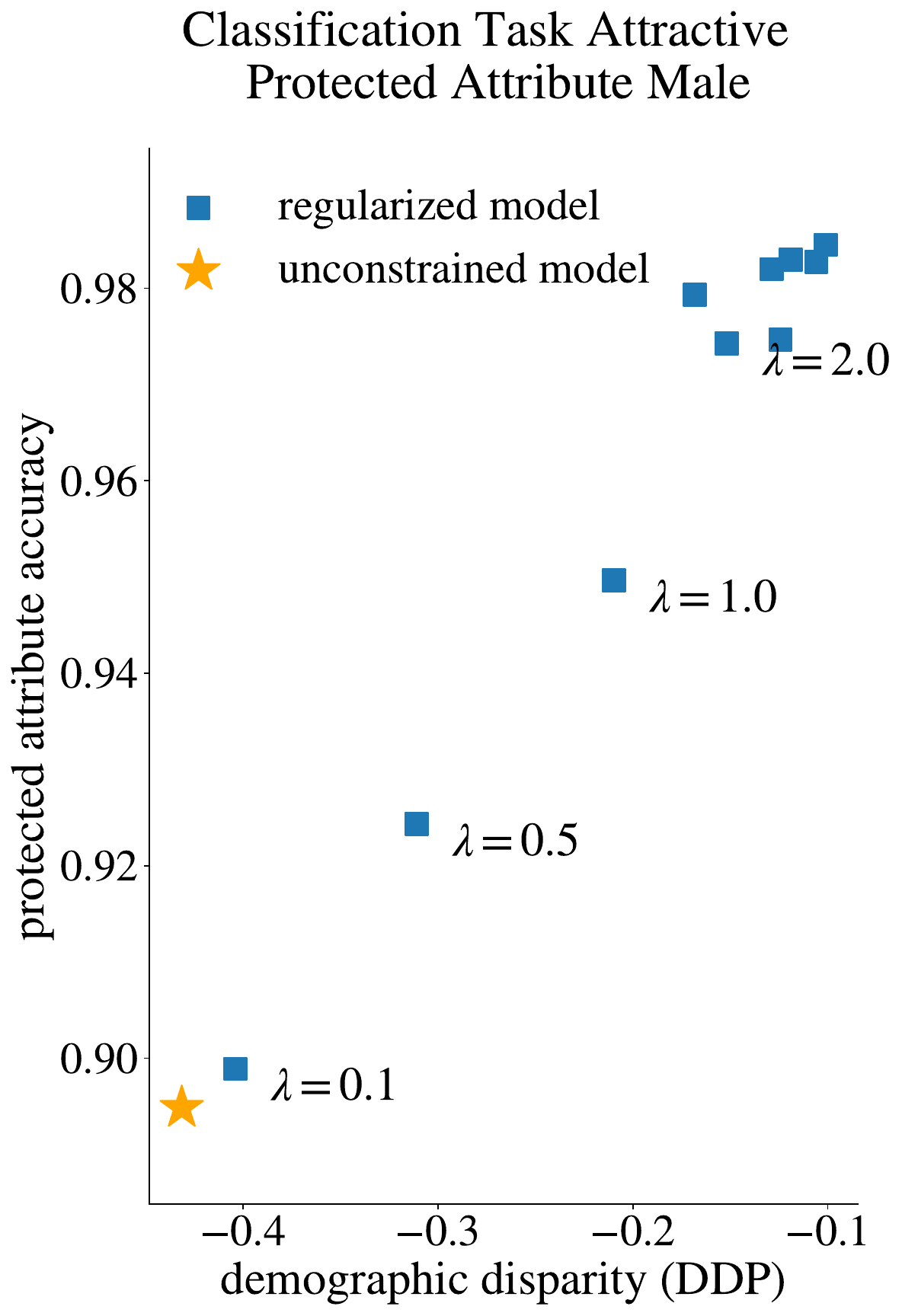}%
\caption{\emph{Left:} \textbf{Kendall-tau rank correlations between fairness parameter $\lambda$ and the accuracy of the group classifier} For each pair of protected and target attribute, we first train $12$ regularized ResNet50 models with varying fairness parameter~$\lambda$. Subsequently, the linear group classifier is trained on the frozen last-layer representation to predict the protected~attribute. From the resulting $12$ pairs of $\lambda$ and protected attribute accuracy, we test for a monotonic relationship between $\lambda$ and protected attribute accuracy by computing the Kendall-tau rank correlation. The color and size of a square  correspond to the value and magnitude of the correlation coefficient. \textbf{For almost all tasks, the accuracy of predicting the protected attribute increases as the fairness parameter increases.} \emph{Right:} We show how the linear separability of gender increases with the regularization strength when the target attribute is \textsc{Attractive} and the protected attribute is \textsc{Male}. As we increase the fairness parameter $\lambda$, the group classifier accuracy increases by up to more than $8\%$, reaching 98\% accuracy.}\label{fig:correlation_table}
\end{figure}

Here, we provide evidence that deep networks that use  a fairness regularizer or preprocessing to satisfy demographic parity,  learn an internal representation that separates groups, thus allowing each group to be treated differently.
To measure how well a fair network separates the protected groups, we examine how accurately a linear classifier can recover the protected attribute from the output of the last layer. We present results for the fairness regularizer~$\widehat{\mathcal{R}}_{\mathrm{DP}}(f)$ defined in \eqref{def_fairness_regularizer} on one dataset;  see Appendix~\ref{suppsec:implicit_awareness} for the results with \emph{Massaging}~\cite{Kamiran2012}, another regularizer and on a second dataset\footnote{Code is available at \url{https://github.com/mlohaus/disparatetreatment}.}.

\textbf{Experimental Setup} From the CelebA image dataset \cite{liu2015faceattributes} and nine of its 40 binary attributes, we choose a target attribute and a protected attribute, such as \textsc{Smiling} and \textsc{Young}. We use \cite{ramaswamy2020gandebiasing} as reference for restricting our study to only nine of the 40 attributes. For each distinct pair of target and protected attribute, we train ResNet50 models for twelve different values of the fairness parameter~$\lambda$. For each model, we then train a linear classifier using logistic regression to predict the protected attribute from the model's frozen last-layer representation. We refer to these classifiers as the group classifiers. See Appendix~\ref{suppsec:datasets_details} for technical details and descriptions of the datasets. 

\textbf{Evaluation} For each pair of target and protected attribute, we evaluate if increasing $\lambda$ increases the accuracy of the group classifier. We test for a monotonic relationship using the Kendall-tau correlation~$\tau$ \citep{kendall1945}  on $12$ datapoint pairs (consisting of  fairness parameter~$\lambda$ and the accuracy of the group classifier). Additionally, we compute a two-sided p-value for the null hypothesis of independence between $\lambda$ and the group classifier's accuracy. Since the regularized approach can collapse to a trivial near-constant classifier  when $\lambda$ is too large, we discard models when their accuracy is too close to the one of the constant classifier (concretely, if their accuracy is not at least the constant classifier’s accuracy plus 25\% of the additional accuracy that the unconstrained classifier achieves).

\textbf{Results} Figure~\ref{fig:correlation_table} summarizes our results. For $71$ out of $72$ experiments the Kendall-tau correlation shows a positive correlation between $\lambda$ and the protected attribute accuracy, with $p<0.05$ for $62$ experiments. Those experiments show a very strong monotonic relationship with a correlation higher than $0.49$. For nine experiments, we observe lower but still positive values of the correlation coefficient~$\tau$. Only for one experiment we observe  $\tau<0$, but with insignificant $p=0.64$. 

\textbf{Conclusion} For both fairness regularizers and the \emph{Massaging} preprocessing method we find that \textbf{increasing the fairness parameter increases the ability to recover the protected attribute from the last layer.} This increase in accuracy is a cause for concern since disparate treatment can occur if a system is able to infer the protected attribute~(cf.~Section~\ref{sec:preliminaries}). We build on this initial analysis in Sections~\ref{sec:explicit_approach} and~\ref{sec:equivalence_and_disparate_treatment} and show that disparate treatment occurs in practice.

\begin{figure}
    \centering
    \includegraphics[width=\textwidth]{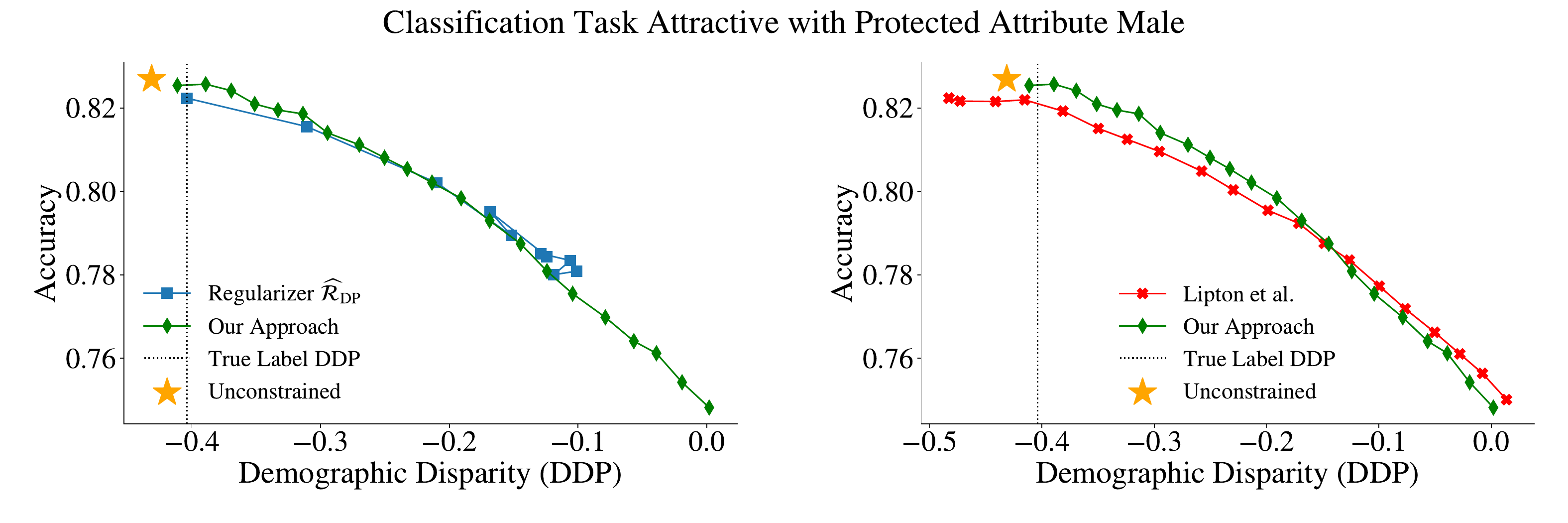}
    \caption{\textbf{Comparison of different single architecture approaches} We compare our group aware model to a fairness-regularized model (\emph{left plot}) and the approach of \citet{lipton2018} (\emph{right plot}) on the task of predicting the target \textsc{Attractive} with respect to the protected attribute  \textsc{Male}.
    For all methods, we observe the typical trade-off:  as the model becomes fairer (DDP is closer to 0), the target accuracy for \textsc{Attractive} decreases. All methods obtain  similar accuracy for a particular DDP value. However, the regularizer approach is unable to achieve near perfect fairness and saturates around a DDP value of~$-0.1$. Note that \citet{lipton2018} requires the protected attribute at test time, while we infer the protected attribute. More datapoints are shown in the scatter plot for \citet{lipton2018} and our approach because they can be generated by varying thresholds without retraining. In contrast, the regularized approach requires a full-retraining for every choice of fairness parameter.
    }
    \label{fig:fairness_vs_accuracy}
\end{figure}

\section{Explicit Group Awareness for Enforcing Demographic Parity}\label{sec:explicit_approach}
In this section, we evaluate a novel post-processing approach \textbf{explicitly designed to exhibit disparate treatment} while using the same single network architecture used by preprocessing and regularized approaches. In short, a network with two fully-connected output heads on the last-layer representation is trained: one head $f:\mathcal{X}\rightarrow \R$ is trained to minimize a logistic loss over the target variable~$y$, and the other head $g:\mathcal{X}\rightarrow \R$  to minimize a squared loss over the protected attribute~$s$. We use the squared loss to encourage $g$ to take values close to zero or one. Overall, we minimize the following objective:
\begin{align*}
\widehat{\mathrm{L}}(f, g) = \frac{1}{n}\left(\sum_{i=1}^n \widehat{\mathrm{L}}_{\mathrm{BCE}}(\sigma(f(x_i)),y_i) + (g(x_i)-s_i)^2\right).
\end{align*}
A weighted sum of the two heads results in a classifier $F$ that satisfies (approximate) demographic parity. The function $F$ takes the form $F(x) =  f(x) + a_1 g(x) + a_2$ for coefficients $a_1, a_2 \in \R$. Even though we train two heads, we can compress them into a single head and generate \emph{a single-headed architecture that makes the same decisions}. More precisely if $f(x)=w_f\cdot z(x)+b_f$ and $g(x)=w_g\cdot z(x)+b_g$, with $z(x)$ being the last-layer representation, then $F(x)=(w_f+a_1  w_g)\cdot z(x)+(b_f+a_1 b_g +a_2)$. This results in a single-headed network with identical architecture to the preprocessing and regularized approaches. We consider our approach a post-processing method as it doesn't induce a fair classifier until we compress the two heads.

As we show, there are substantial advantages to post-processing, both in terms of stability and efficiency of training. However, the key limitation of existing methods is that they either require knowledge of the protected attribute \cite{lipton2018, alabdulmohsin2021near} or more complex systems at evaluation time that must also infer the protected attribute \cite{woodworth2017, oneto2019}. We remove this limitation, by compressing our two-headed approach into the same architecture used by the preprocessing and regularized approaches. This matching architecture is key to the legal argument in the following sections: our approach explicitly exhibits disparate treatment while sharing a common architecture and making similar decisions to other approaches.

\textbf{Model Construction} We consider a standard network backbone (e.g., ResNet50), with two heads. We present results with a ResNet50 in the body of the paper, and include further results with the MobileNetV3-Small architecture in Appendix~\ref{suppsec:explicit_model}. To find $a_1$ and $a_2$ such that the predictions of the thresholding rule $\indic(F(x) > 0)$ are fair and maximally accurate, we perform a grid search on validation data and select the most accurate classifier that does not exceed a given demographic~disparity. 

We compare our approach to \citet{lipton2018} who's similar approach provides optimal per-group thresholds, but requires explicit knowledge of the protected attributes at test time. If the group classifier~$g$ is perfect, our approach and Lipton's approach coincide. If $g$ does not predict the protected attribute well, our procedure can only perform worse than Lipton's (cf. Theorem~4 in their paper). In our computer vision setting, however, the accuracy of predicting the protected attribute is typically very high (e.g., on \textsc{Male} from CelebA, we achieve an accuracy of around $98\%$), and, as we show, the performance of our approach is very close to that of \cite{lipton2018}.
 
\def\rot{\rotatebox}
\newcommand{\Cross}{\mathbin{\tikz [x=1.9ex,y=1.9ex,line width=.3ex] \draw (0,0) -- (1,1) (0,1) -- (1,0);}}
 \begin{table}[t]
        \caption{\looseness=-1 \textbf{Accuracy of single architecture approaches under strict fairness constraints.} We show the accuracy of various approaches at substantially decreased demographic disparity. In the first block, we require the DDP to be half  that of the DDP of the unconstrained classifier, in absolute value. For all methods, we report the most accurate model of all sufficiently fair models. In the second, the DDP can be at most $20$\%.
        Failure to reach the required fairness is indicated with a cross. For small reductions in disparity, all methods have a similar accuracy-fairness trade-off. However, the regularized approach often fails to find sufficiently fair solutions. 
        \textbf{Our approach always finds a sufficiently fair solution} and is comparable to Lipton's approach, which requires the protected attribute at test time.\label{tab:two_heads_comparison}}

        \centering
        \resizebox{\columnwidth}{!}{
        \begin{tabular}{@{}c c@{}c@{}c@{}c@{}c@{}c@{}c@{}c@{}}&\rot{35}{Attractive}&\rot{35}{Bags\_Under\_Eyes}&\rot{35}{Bangs}&\rot{35}{Black\_Hair}&\rot{35}{Eyeglasses}&\rot{35}{High\_Cheekbones}&\rot{35}{Smiling}&\rot{35}{Young}\\\toprule
        &&&\multicolumn{3}{c}{50\% disparity reduction} \\\midrule
        Lipton& 0.7955& 0.8452& 0.9469& 0.8959& 0.9680& 0.8641& 0.9240& 0.8770\\Our Approach& 0.8021& 0.8419& 0.9463& 0.8905& 0.9775& 0.8615& 0.9227& 0.8756\\Massaging& 0.7992& 0.8337& 0.9481& 0.9002& 0.9718& 0.8546& 0.9207& 0.8679\\Regularizer $\widehat{\mathcal{R}}_{\mathrm{DP}}$& 0.8021& 0.8274& $\Cross$& 0.9023& $\Cross$& 0.8350& $\Cross$& 0.8701\\ \midrule
&&&\multicolumn{3}{c}{80\% disparity reduction}\\\midrule
            Lipton& 0.7719& 0.8344& 0.9336& 0.8959& 0.9632& 0.8456& 0.9137& 0.8617\\Our Approach& 0.7698& 0.8332& 0.9301& 0.8881& 0.9647& 0.8436& 0.9118& 0.8592\\Massaging& 0.7674& 0.8185& $\Cross$& 0.8989& $\Cross$& $\Cross$& $\Cross$& 0.8573\\Regularizer $\widehat{\mathcal{R}}_{\mathrm{DP}}$& $\Cross$& 0.8274& $\Cross$& $\Cross$& $\Cross$& $\Cross$& $\Cross$& $\Cross$
        \end{tabular}
        }
    \end{table}

\textbf{Results} Figure~\ref{fig:fairness_vs_accuracy} compares the accuracy-fairness trade-off of our approach, the regularizer approach, and~\citet{lipton2018}. Since the computational cost of fully training $12$ regularized models (with the same $12$ fairness parameter values as in Section~\ref{sec:implicit_awareness}) is much higher than training one model for our approach and Lipton's, we compute $20$ solutions with our grid search or Lipton's greedy search. All three methods offer similar accuracy for a particular level of demographic disparity. However, for the regularized model, it is difficult to control this trade-off or to find very fair solutions. In contrast, our approach and \cite{lipton2018}  allow easy selection of a model with a particular demographic disparity. While the performance of our approach and~\cite{lipton2018} is similar,  we do not  require the  protected attribute at test time.

Table~\ref{tab:two_heads_comparison}  reports the accuracy  obtained  under strict fairness constraints. With respect to the accuracy-fairness trade-off all methods perform comparably. However, if we require a substantial reduction of unfairness (second block of rows in Table~\ref{tab:two_heads_comparison}), the regularizer approach and preprocessing  often fail to find a valid solution. In contrast,~\citet{lipton2018} and our  approach always find sufficiently fair solutions due to their direct search for per-group thresholds.

\textbf{Conclusion} Our method has several advantages: (i) Other approaches require training a new model for every fairness parameter $\lambda$, which might make tuning $\lambda$ very expensive until a desirable level of fairness is reached. Our approach only requires a single explicit model and the output scores of the two heads. (ii) Due to the simple weighted sum, we make the influence of the group classifier in the final decision explicit and transparent. In summary, \textbf{our approach reliably finds high accuracy solutions for a given demographic disparity} without requiring the protected attribute at test time.

\section{Disparate Treatment in Fair Networks}\label{sec:equivalence_and_disparate_treatment}
In this section we examine the tight relationship between our explicit approach and the behavior of fair neural networks.   
First, we reconstruct the fair network with our group-aware method by building a weighted sum of the two heads and merging them into one (Section~\ref{sec:explicit_approach}). Second, from a given fair neural network, we recover the corresponding unconstrained model using only the fair network and the group classifier from our approach. The relationship between fair model and our approach allows us to identify individuals who are treated differently based on inferred group membership and to demonstrate disparate treatment.

\subsection{Fair Networks Behave like the Explicit Approach}\label{sec:model_equivalence}

\begin{figure}
    \centering
    \includegraphics[width=0.4\textwidth]{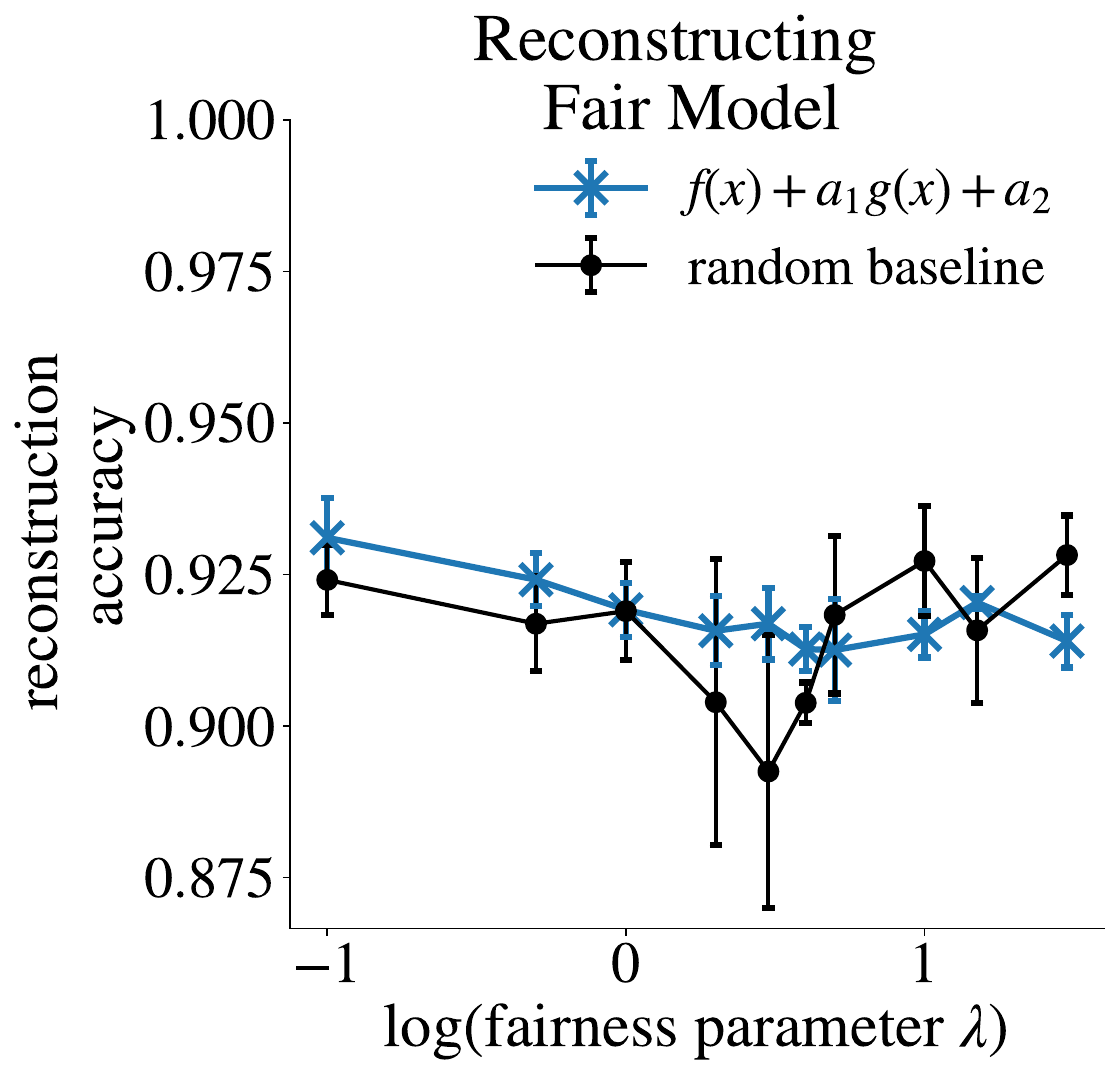}\hspace{1cm}
    \includegraphics[width=0.4\textwidth]{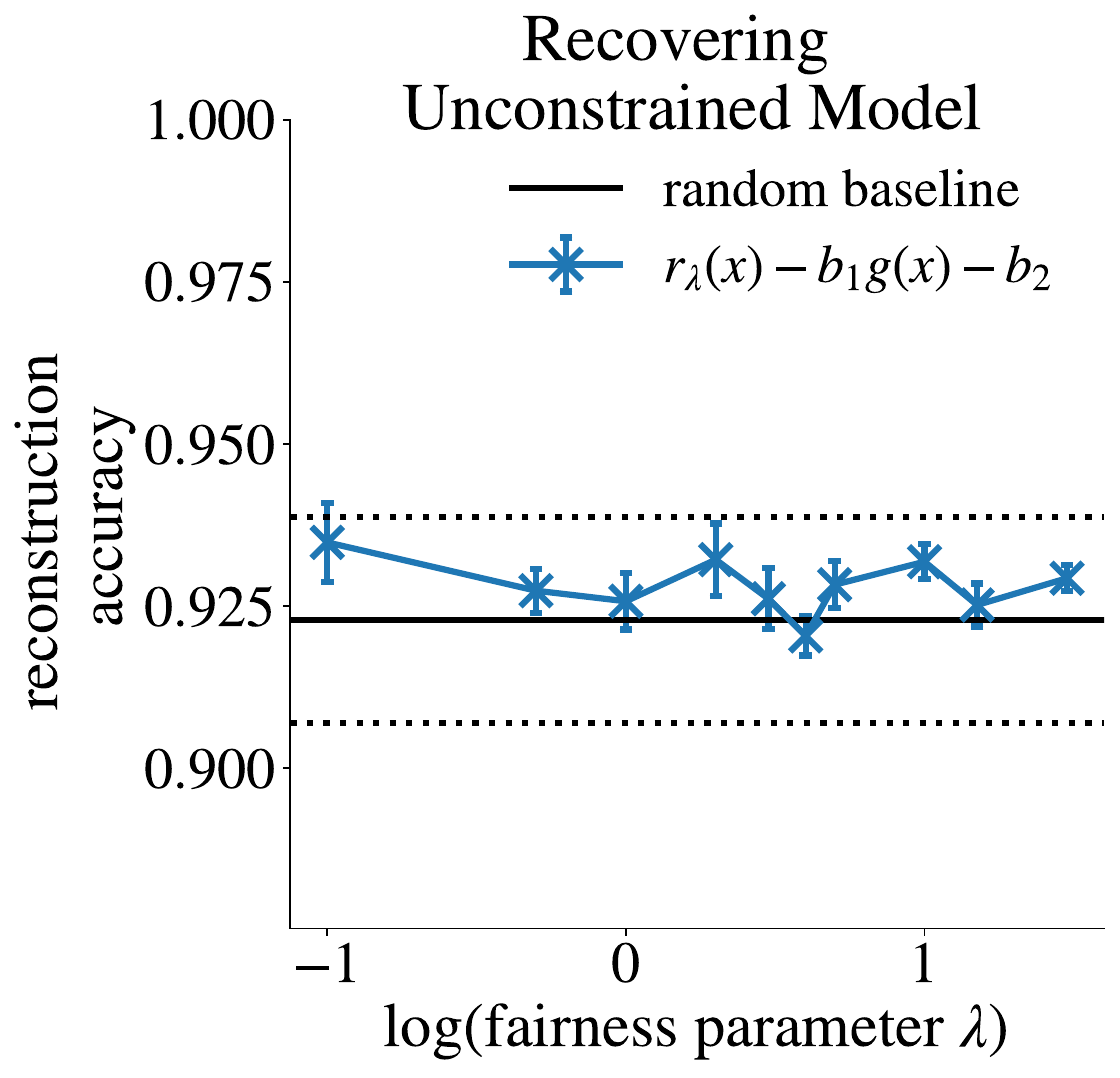}
    \caption{\looseness=-1\emph{Left:} \textbf{Reconstructing the fair classifier.} For a range of  $\lambda$ ($x$-axis) we find $a_1$ and $a_2$ so that $f + a_1 g + a_2$  mimics the predictions of a regularized classifier $r_\lambda$. For the entire range of fairness parameters, the predictions of the regularized model are closely recovered by our approach.
    \emph{Right:}  \textbf{Recovering the unconstrained classifier.}  We find parameters $b_1$ and $b_2$ such that $r_\lambda- b_1 g - b_2$ recovers the predictions of an unconstrained classifier $h$.  From the fair model $r_\lambda$ and the group classifier $g$ we can replicate an unconstrained classifier's predictions, as accurately as another unconstrained model. Here, we predict \textsc{Attractive} with protected attribute \textsc{Male}. Reconstruction accuracy is directly comparable to rerunning the method with a different random seed (black lines, random baseline).
    }\label{fig:approx_to_reg_models}
\end{figure}

\looseness=-1 We recover the predictions of a fair model $r_{\lambda}$ by using the target task head~$f$ and group classifier~$g$. We use logistic regression to find parameters $a_1, a_2 \in \R$ such that $\indic(f(x) + a_1 g(x) + a_2 > 0)$ accurately replicates $\indic(r_{\lambda}(x) > 0)$. Since we can merge the weighted sum into a single output head, our two-headed approach has thus found a model of the same original architecture with one output head that predicts equivalently to the fair model. Next, we show that it is possible to recover the predictions of the unconstrained model~$h$ with a weighted sum of the fair classifier~$r_{\lambda}$ and the group classifier~$g$. We run logistic regression to find $b_1, b_2 \in \R$ such that $\indic(r_{\lambda}(x) - b_1 g(x) - b_2 > 0)$ recovers the prediction $y = \indic(h(x) > 0)$. Coefficients $a_1$ and $a_2$, or $b_1$ and $b_2$ are found using  validation data. 

\textbf{Error Bars and Random Baseline} For both experiments, a substantial challenge lies in the random behavior of training a deep network. The solution found highly depends on its initial seed.
To take this instability into account, we provide baselines that measure how  decisions vary when retraining networks. In the first experiment, we compute error bars by repeating the process for five different initial random seeds of $r_{\lambda}$. We compute a random baseline by comparing the five models to a sixth, differently initialized model $r_{\lambda}$. In the second experiment, we recover five unconstrained models trained with different initial seeds (for the error bars) and we compute a random baseline by comparing the predictions to a sixth unconstrained model.

\textbf{Results} In Figure~\ref{fig:approx_to_reg_models} we show regularized ResNet50 models trained to predict \textsc{Attractive}; the protected attribute is \textsc{Male}~(see Appendix~\ref{suppsec:model_equivalence} for other tasks, models, and preprocessing results). The left panel evaluates how accurately our explicit approach recovers the predictions of the regularized model~$r_{\lambda}$. We find that most of the error in recovering predictions can be attributed to classifier instability, and that retraining a classifier from scratch with a new random seed gives similar disagreement to using our reconstructed classifier. In the right panel of Figure~\ref{fig:approx_to_reg_models}, we plot the reconstruction accuracy of $r_{\lambda} - b_1 g - b_2$ with respect to the unconstrained classifier. By simply adding the group classifier response $g(x)$ to $r_{\lambda}(x)$, we obtain the predictions of the unconstrained classifier. {\bf Compared to the baseline, we recover the unconstrained classifier responses for all $r_{\lambda}$ with similar fidelity to simply retraining the target classifier from scratch.} 
    
\subsection{Identifying Disparate Treatment in Deep Networks}\label{sec:disparate_treatment}

We can now quantify the disparate treatment of a neural network. By exploiting our explicit estimation of the protected attribute, we can ask the counterfactual question: how would the decision have changed if the individual had belonged to the other group?

\textbf{Experimental Setup} As described in the previous subsection, we find the closest weighted sum $f + a_1g + a_2$ of the two heads that best replicates the decisions of a given model $r_\lambda$. Then, for every individual $x$ in the test set, we replace the group classifier response $g(x)$ with the median output of the group that $x$ does \emph{not} belong to. We evaluate how many times the prediction changes when the second head output is replaced by this counterfactual.

\begin{figure}
    \centering
    \subcaptionbox{ResNet50 with regularizer $\widehat{\mathcal{R}}_{\mathrm{DP}}$ predicting \textsc{Attractive}. }{\includegraphics[width=0.31\textwidth]{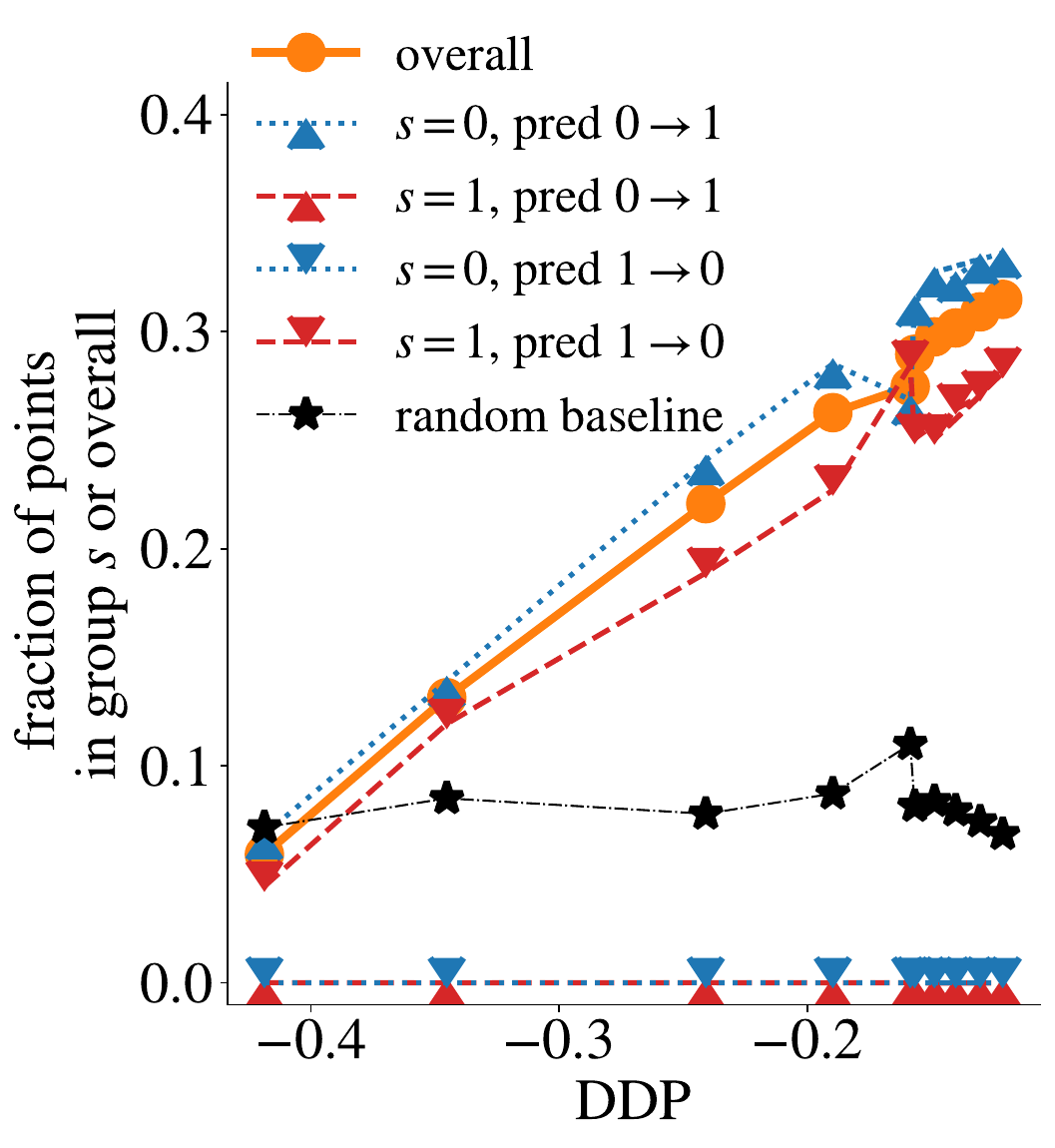}}\hfill
    \subcaptionbox{ResNet50 with regularizer $\widehat{\mathcal{R}}_{\mathrm{DP}}$  predicting target attribute \textsc{Smiling}. }{\includegraphics[width=0.33\textwidth]{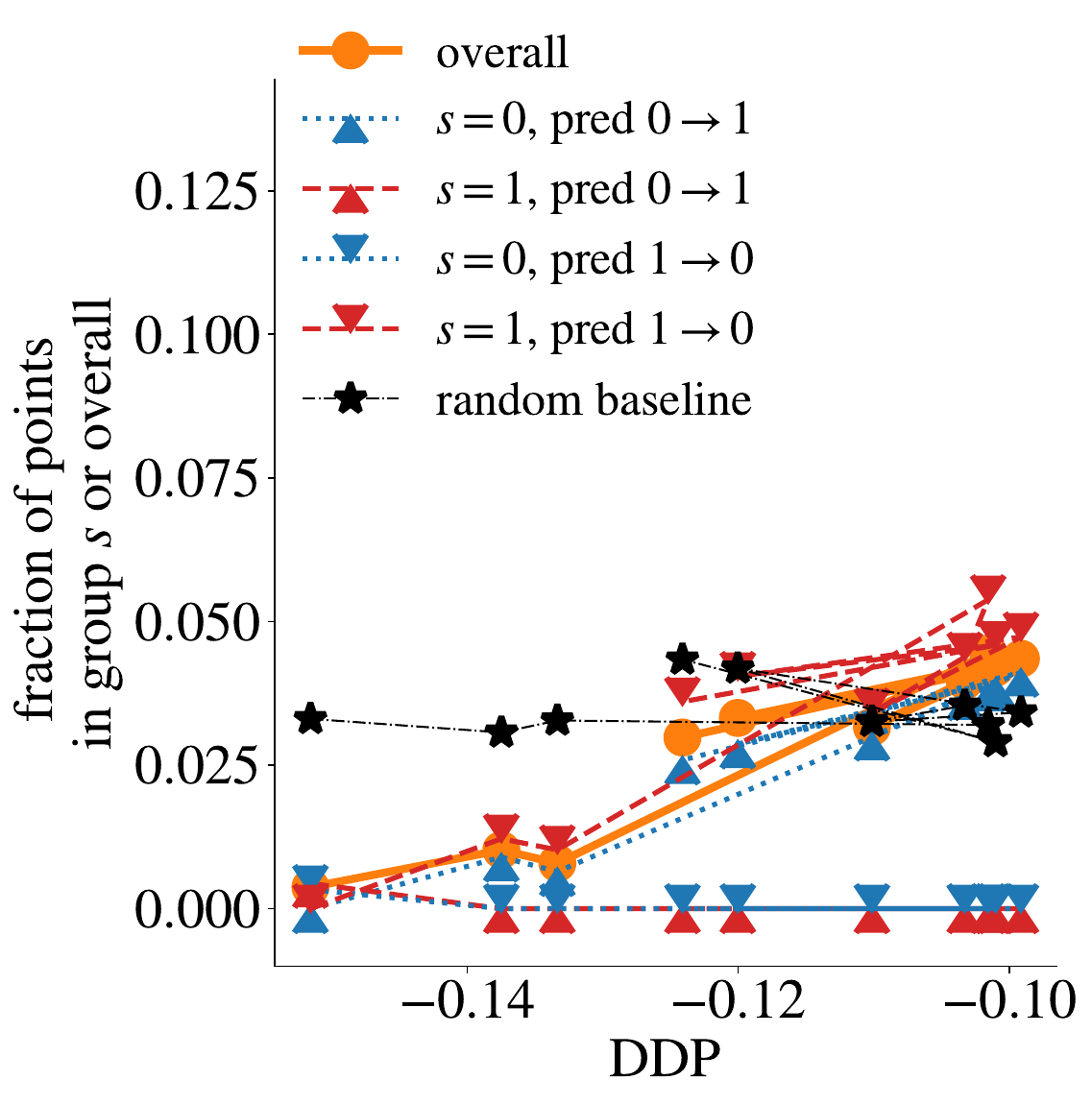}}\hfill
    \subcaptionbox{MobileNetV3-Small with Massaging predicting \textsc{Attractive}. }{\includegraphics[width=0.32\textwidth]{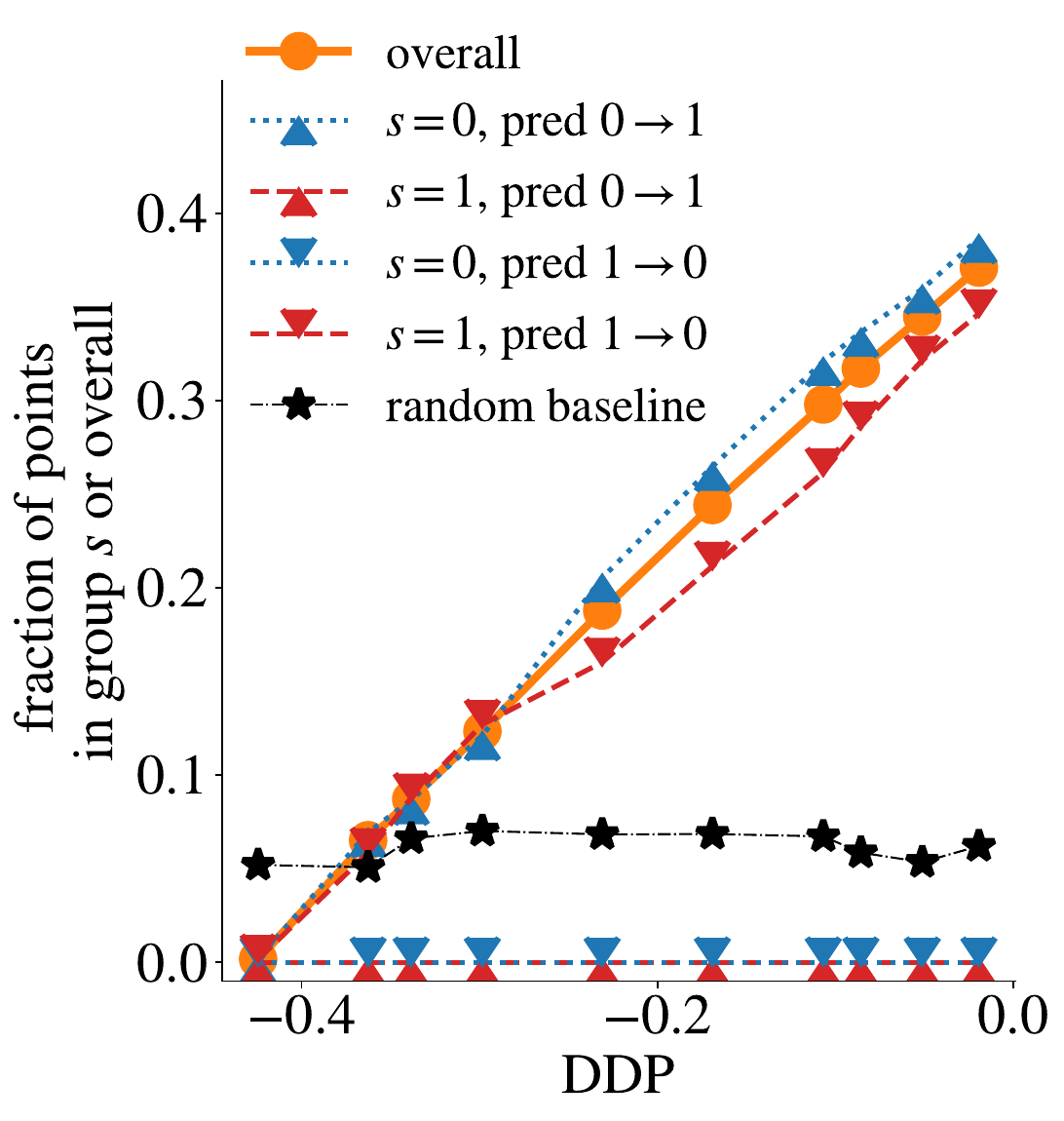}}
    \caption{\looseness=-1 \textbf{Uncovering disparate treatment.} How many individuals are treated differently in fair neural networks based on their protected attribute? With our approach, we find a single-headed network that approximates the decisions of the fair model (c.f. Section~\ref{sec:model_equivalence}). Our explicit formulation allows us to replace the group classifier output $g(x)$ of an individual $x$ from group $s \in \{0, 1\}$ with the median output $\bar{g}_{1-s}$ of the other group $1-s$. We plot the proportion of all points where the label changes (orange curves), and the proportion of points in each protected group (red and blue) for which the prediction either changed from $0$ to $1$ (markers pointing up) or changed from $1$ to $0$ (markers pointing down). \emph{Left:} As the fairness parameter increases and fairness of the regularized model improves (zero is fair), the proportion of changed predictions increases. For the fairest model, around $30\%$ of points would obtain a different outcome if their perceived gender changed. \emph{Center:} When we observe a small change in demographic disparity, we observe only a small proportion of points treated differently. In this plot the DDP is improved from $-0.15$ to $-0.1$. In these cases it is difficult to determine if disparate treatment is occurring. \emph{Right:}  We conduct the same experiment for \emph{Massaging}~\cite{Kamiran2012}. Again, up to $35\%$ of predictions change if the predicted group membership~changes.
    }\label{fig:disparate_treatment_explicit_approach}
\end{figure}

\textbf{Results}  Figure~\ref{fig:disparate_treatment_explicit_approach} shows the proportion of individuals for whom their prediction changes. For the fairest models in the left panel, up to $30$\% of all individuals receive a different outcome when their second head output $g(x)$ is replaced by the median output of the other group. This is substantially more than the number of changed predictions, which we obtain when retraining with a different random seed (roughly $7$\% of the points). As expected, the proportion of changed predictions linearly increases with model fairness (governed by the parameter~$\lambda$). Similar behavior occurs for both the regularized approach (Figure~\ref{fig:disparate_treatment_explicit_approach}, left) and preprocessing (right).

While the behavior of our  system is difficult to distinguish from that of a retrained fair classifier,  the  disagreement between retrained classifiers means that we cannot point to an individual and conclude that they received a different decision because of their protected attribute.  Nonetheless, in scenarios where changing the protected attribute alters a much greater proportion  of decisions 
than the proportion of  decisions where the classifiers disagree (see Figure \ref{fig:disparate_treatment_explicit_approach} left and right, in contrast to center) we can conclude that it is likely particular individuals  suffered disparate treatment.

\textbf{Conclusion}  When fair networks show the same behavior as our explicit awareness model, we can analyze the influence of group membership. Using our explicit approach, we can evaluate how fair networks systematically treat individuals differently on the basis of their protected attribute. 

\section{Legal Implications of our Analysis}\label{sec:legal_main_body}
We provide an extensive discussion of the legal implications of our analysis in Appendix~\ref{sec:legal}; here we  provide an overview of our argument.

The analysis set out in the appendix is restricted to areas where the doctrine of disparate treatment is relevant. This includes areas where decisions are made concerning an individual's access to: education, employment, and housing. We start by noting that by design our  approach exhibits disparate treatment. It assigns individuals into different racial or gender-based groups and uses this to alter cutoff scores in a way that explicitly violates Title VII \cite{civilrights_title7} as outlined in the introduction. What is less straightforward is the relationship of the methods that we have shown to have the same systematic behavior as our new approach.   

The overall legal argument can be decomposed into three parts:   

\begin{enumerate}
    \item Disparate treatment may occur even if the treatment is based on inferred attributes (such as race or sex) rather than explicitly provided attributes.
    \item By construction, our explicit formulation (Section~\ref{sec:explicit_approach}) exhibits disparate treatment. 
    \item Other fairness approaches considered exhibit the same behavior as our explicit approach, and as the relevant tests for disparate treatment are based on systematic behavior, these approaches also exhibit disparate treatment. 
\end{enumerate}
Finally, we  explicitly identify the individuals likely disadvantaged by enforcing fairness, and  consider circumstances where the use of such systems may be acceptable, and look at existing legal arguments.

\section{Conclusion}
\looseness=-1 We showed how popular existing methods for enforcing demographic parity in deep networks learn an internal representation that is more predictive of the protected attribute. The close coupling between the behavior of these approaches to our explicit model provides a tool to identify individuals who are likely to be systematically favored or disfavored by virtue of their protected attribute and to conclude that existing methods for enforcing fairness do so through disparate treatment. We have shown that the use of \emph{some} fairness methods on \emph{some} datasets exhibits disparate treatment. Our findings do not imply  that \emph{all} methods for enforcing demographic parity suffer from disparate treatment, merely that some can, and that caution should be used when deploying such methods. 

Furthermore, our approach for enforcing fairness offers a more efficient and reliable way of enforcing a chosen degree of demographic parity. Unlike regularization-based approaches that require multiple training runs with different regularization parameters to find a desired trade-off, our approach allows for joint training of both heads once, and then a search for the desired trade-off by tuning weights using precomputed network responses on a validation set, and then finally merging the two heads. As such, it is possible to efficiently find a family of classifiers from the original architecture, of varying fairness and accuracy for little more compute than training an unfair classifier in the first place.

In hindsight, our findings are perhaps unsurprising. Regularizing methods try to balance decisions made by the classifier while minimizing the logistic loss. This requires  the calibration of the classifier to be preserved as much as possible. To this end, a fairly regularized classifier should maximally distort points closest to the decision boundary, while trying to minimally alter the scores of other points. Massaging methods train one classifier, and then push points closest to the decision boundary over it, before training a new classifier to replicate these decisions. While both methods explicitly treat members of different groups differently, they do so at training time and rely on this behavior generalizing to unseen data. In contrast, our approach simply moves the decision boundary at test-time for points identified as belonging to a particular group. However, following these intuitions, it is not unreasonable that all three methods relabel the same set of points that an unconstrained classifier would be least confident about.

While our analysis presents several challenges to deploying fair ML systems in the US, it is consistent with other rulings on discrimination in US law. In general, the US requires that considerations of equity and affirmative action are satisfied by shaping an entire process to be more inclusive, and not simply by imposing race or gender-based quotas on outcomes \cite{joshi2018racial, fazelpour2020algorithmic}. However, the opaque nature of ML makes it extremely challenging to define fair algorithms without formulating the definitions in terms of outcomes\footnote{See \cite{kim2022race} for an excellent legal analysis that includes the legality of other approaches that do not explicitly rely on outcomes.}. For this reason, we believe that legal reform is needed to explicitly allow the use of fair ML techniques as a tool to reduce disparate impact and increase equity.


\begin{ack}
Chris Russell is also a member of the Governance of Emerging Technologies and the Trustworthy Auditing for AI project at the Oxford Internet Institute. We are grateful to other members, particularly Sandra Wachter and Johann Laux for their advice. The legal discussion strongly benefited from their thoughtful suggestions. Michael Lohaus is a member of the International Max Planck Research School for Intelligent Systems (IMPRS-IS). We thank Dominik Zietlow for his practical advice.
\end{ack}

\newpage

{
\small
\bibliographystyle{plainnat}
\bibliography{ref}
}

\section*{Checklist}

\begin{enumerate}

\item For all authors...
\begin{enumerate}
  \item Do the main claims made in the abstract and introduction accurately reflect the paper's contributions and scope?
    \answerYes{}
  \item Did you describe the limitations of your work?
    \answerYes{In the conclusion section we emphasize that we have only found disparate treatment for some fairness methods on some datasets.}
  \item Did you discuss any potential negative societal impacts of your work?
    \answerYes{Our analysis has legal implications which we discuss shortly in Section~\ref{sec:legal_main_body} and in more depth in Appendix~\ref{sec:legal}.}
  \item Have you read the ethics review guidelines and ensured that your paper conforms to them?
    \answerYes{}
\end{enumerate}

\item If you are including theoretical results...
\begin{enumerate}
  \item Did you state the full set of assumptions of all theoretical results?
    \answerNA{}
        \item Did you include complete proofs of all theoretical results?
    \answerNA{}
\end{enumerate}

\item If you ran experiments...
\begin{enumerate}
  \item Did you include the code, data, and instructions needed to reproduce the main experimental results (either in the supplemental material or as a URL)?
    \answerYes{We provide code examples as we simply train PyTorch implementations of popular neural networks. Code is available at \url{https://github.com/mlohaus/disparatetreatment}. We provide all details on the data and experimental details in Appendix~\ref{suppsec:technical_details}.}
  \item Did you specify all the training details (e.g., data splits, hyperparameters, how they were chosen)?
    \answerYes{See answer above.}
        \item Did you report error bars (e.g., with respect to the random seed after running experiments multiple times)?
    \answerYes{}
        \item Did you include the total amount of compute and the type of resources used (e.g., type of GPUs, internal cluster, or cloud provider)?
    \answerNo{}
\end{enumerate}

\item If you are using existing assets (e.g., code, data, models) or curating/releasing new assets...
\begin{enumerate}
  \item If your work uses existing assets, did you cite the creators?
    \answerYes{We use pretrained PyTorch models and publicly available data as discussed in Appendix~\ref{suppsec:datasets_details}.}
  \item Did you mention the license of the assets?
    \answerYes{ }
  \item Did you include any new assets either in the supplemental material or as a URL?
    \answerNo{}
  \item Did you discuss whether and how consent was obtained from people whose data you're using/curating?
    \answerNo{}
  \item Did you discuss whether the data you are using/curating contains personally identifiable information or offensive content?
    \answerNo{We use standard image datasets.}
\end{enumerate}

\item If you used crowdsourcing or conducted research with human subjects...
\begin{enumerate}
  \item Did you include the full text of instructions given to participants and screenshots, if applicable?
    \answerNA{}
  \item Did you describe any potential participant risks, with links to Institutional Review Board (IRB) approvals, if applicable?
    \answerNA{}
  \item Did you include the estimated hourly wage paid to participants and the total amount spent on participant compensation?
    \answerNA{}
\end{enumerate}

\end{enumerate}


\newpage

\appendix

\section*{Appendix}

\section{Legal Implications of our Analysis}\label{sec:legal}
The analysis set out in this section is restricted to areas where the doctrine of disparate treatment is relevant. This includes areas where decisions are made concerning an individual's access to: education, employment, and housing.
We start by noting that by design our  approach exhibits disparate treatment. It assigns individuals into different racial or gender-based groups and uses this to alter cutoff scores in a way that explicitly violates Title VII \cite{civilrights_title7} as outlined in the introduction. What is less straightforward is the relationship of the methods that we have shown to have the same systematic behavior as our new approach.

\subsection*{Overview} Our  argument can be decomposed into three parts. We address each point in detail below:
\begin{enumerate}
    \item Disparate treatment may occur even if the treatment is based on inferred attributes (such as race or sex) rather than explicitly provided attributes.
    \item Our explicit formulation (Section~\ref{sec:explicit_approach}) exhibits the behavior used to \emph{indirectly} identify disparate treatment.
    \item Other fairness approaches considered exhibit the same behavior as our explicit approach, and as the relevant tests for disparate treatment are based on systematic behavior, these approaches also exhibit disparate treatment. 
\end{enumerate}
Finally, we  explicitly identify the individuals likely disadvantaged by enforcing fairness, and  consider circumstances where the use of such systems may be acceptable, and look at existing legal arguments.

\begin{enumerate}
    \item[\textbf{1.}] \textbf{Implicit Disparate Treatment}
Multiple machine learning papers \citep{zafar2017fairness, donini2018, perrone2021fairbayesian, agarwal_reductions_approach} have asserted that machine learning systems that do not explicitly take into account knowledge of the protected attribute at prediction time cannot be performing disparate treatment. From a legal perspective, this is an oversimplification. Many of the legal tests for disparate treatment involve a demonstration of intent to treat protected groups differently \citep{king2020blurred}, and it is irrelevant if knowledge of the groups is given as data from a trusted party or inferred from a photograph or other data. As an example of case law supporting this, \cite{hellman2020measuring} gives Hunt v.\ Cromartie \cite{1999hunt} where the plaintiffs demonstrated that location was used as a proxy for race in an instance of disparate treatment. 
The situation considered here is even more extreme than that of Hunt v.\ Cromartie. As we use photographic data as input, to deny that disparate treatment can occur
here is the same as denying that disparate treatment can occur on the basis of
an individual's appearance. 
\item[\textbf{2.}] \textbf{The Disparate Treatment of Our Approach}
Systems which explicitly alter scoring on the basis of race or gender are widely acknowledged as being examples of disparate treatment, with \cite{hellman2020measuring} writing that there is such wide agreement that it is not worth discussing.
While creating a system that makes explicit use of a protected attribute when making decisions demonstrates intent, it is not the only way to do so.  In particular, as it is difficult to explicitly demonstrate intent when someone is either unable or unwilling to explain honestly why they made decisions, the courts recognize indirect evidence of the form:
``. . . evidence, whether or not rigorously statistical, that employees similarly situated to the plaintiff other than in the characteristic (pregnancy, sex, race, or whatever) on which an employer is forbidden to base a difference in treatment received systematically better treatment'' \citet{Troupevmay} with similar judgments repeating these arguments occurring in \citet{} . By design, our approach explicitly treats `similarly situated' individuals, i.e. those who receive a similar score $f(x)$ from a classifier trained without consideration of their protected attribute, differently by changing their score and adding the term $a_1 g(x) + a_2$ which explicitly depends on their inferred protected attribute.  

\item[3.] \textbf{The Disparate Treatment of Other Approaches}
As the implicit argument of the previous section relies on the systematic behavior of a decision-making system, it can be directly applied to systems trained to satisfy a fairness constraint. In such cases, we only know that the system enforces the constraint, but not necessarily how. However, as we are only concerned with the system behaviour, i.e. the decisions made, if the system is closely mimicked by our approach, the same argument applies, and we can identify those individuals with similar scores $f(x)$ who probably\footnote{As the reconstruction is not exact, we cannot be certain, however, note that the evidence provided does not even need to be \textit{rigorously statistical} \cite{Troupevmay}.} receive different decisions by virtue of their race or gender.
\end{enumerate}

\paragraph{Identifying Disadvantaged Individuals} We can therefore identify individuals that are likely to have received an unfavorable decision by virtue of their inferred protected attribute. As we can recover a close approximation of the decisions of the fair model of the form $r(x)\approx f(x)-a_1 g(x)-a_2$, individuals who initially receive a score $f(x)$ in the region $f(x)\in [a_2,a_1+a_2)$  are likely to receive different decisions by virtue of their inferred race or gender $g(x)$.\footnote{This relies on $g(x)$ being close to an indicator function with $94\%$ of function responses being either $0$ or $1$ with a tolerance of $\pm0.1$.}

\paragraph{Potential Mitigation}
One possible defense, that would only be valid in vary narrow circumstances,  is to reject the relevance of the classifier $f$ trained on ground-truth data without fairness considerations, and to claim that individuals with similar $f(x)$ scores are not in fact ``similarly situated'' \citep{Troupevmay}. It is always possible to generate some classifier\footnote{Here, we consider $f$ an arbitrary classifier, and not necessarily the unconstrained classifier trained on the data.} $f$ from an existing classifier $r$, by subtracting any arbitrary terms of the form $a_1 g(x)+a_2$ from $r(x)$. As such, the existence of $f$ is insufficient to conclude that the existing classifier $r$ exhibits disparate treatment, and we also need to know that individuals with similar scores $f(x)$ are ``similarly situated''. As such, where this unaware classifier $f$ was trained on data that does not correspond to a  direct measure  of performance, and is known to be systematically biased \cite{barocas2016big,wachter2021fairness}, there is little reason to believe that individuals with similar scores are also ``similarly situated''. This argument was proposed outside of algorithmic fairness by \citet{selmi2013indirect} who argued that a stronger defense could have been mounted in \citep{ricci} by challenging the predictive value of the test.  

\subsection*{Existing Legal Arguments}
A full summary of the existing debate is out of scope for what is primarily a machine learning paper, and we only touch upon three papers to indicate the range of opinions. \citet{kroll165accountable} argued that protected attributes should not be used as part of the decision-making system, but that the use of ML fairness methods that  use protected attributes at training time was less likely to be considered an instance of disparate treatment than an ex-post correction that explicitly alters the score of individuals with a particular race or gender. In particular, \citet{kroll165accountable} considered \cite{Troupevmay} and similar judgements and concluded that compared to ex-post measures\footnote{Ex-post methods adjust an already generated scoring system by choosing different thresholds for members of different groups.}  ``incorporating
nondiscrimination in the initial design of algorithms is the safest path that decision makers can take.'' This argument hinges upon an explicit lack of understanding of the behavior of ML systems, i.e., without knowing the details of how an opaque fair system behaves it is not possible to say that it exhibits disparate treatment. \citet{hellman2020measuring} argued that as disparate treatment depends upon intent, in certain circumstances, the use of protected attributes can be acceptable at decision time. \citet{bent2019algorithmic} offered two main arguments: First, that as disparate treatment hinges upon intent, the combined use of racial data (even if only at training time) with any form of fairness constraints shows an expressed intent to treat different races differently, and should also be considered disparate treatment.\footnote{\citet{bent2019algorithmic} argues that this should hold for any form of fairness constraint, not just the forms of demographic parity we consider.}
\footnote{The second argument \citet{bent2019algorithmic} considers is the difference between running an algorithm with and without race-based fairness constraints. \citeauthor{bent2019algorithmic} argues that any individual receiving a change in decision should be considered evidence of a difference in treatment. This argument is problematic due to the non-deterministic behavior of deep learning algorithms. As shown in Figures~\ref{fig:approx_to_reg_models} and \ref{fig:disparate_treatment_explicit_approach}, simply rerunning the same algorithm with a different seed can result in significant changes in labels assigned to the test set, and what is needed is evidence of a systematic change in treatment over and above the expected intrinsic variability. Inherently, such evidence cannot come from considering a single individual, but must occur at the population level. See also related discussion in \cite{kim2022race} regarding the \cite{bent2019algorithmic} and the ill-determined nature of `unfair' classifiers.} 

In response to \citet{bent2019algorithmic}, we note that: \emph{(i)} bias-preserving fairness metrics \cite{wachter2020bias} (i.e., the majority of existing fairness metrics) ensure that classifiers are sufficiently accurate for all groups, by matching the distribution of errors over different groups. \emph{(ii)} \citet{kroll165accountable} also argued that in \cite{ricci} the court's lack of concern regarding the use of race to determine that the scoring mechanism was equally effective for all racial groups indicated that this was a legitimate use of racial data. Putting these two arguments together, it seems likely that enforcing many forms of fairness need not be a form of disparate treatment, and that it depends instead on specific facts of implementation and particularly if it is possible to identify individuals who are systematically disadvantaged by the fact of their race under a fair system.

Our position lies midway between \citet{kroll165accountable} and the first argument of \citet{bent2019algorithmic}, and is simply that a blanket decision if algorithmic fairness violates disparate treatment is inappropriate and depends upon the facts of the particular ML system considered and the data it is deployed on. Our position is that the improved understanding from using the techniques set out in this paper is sufficient to determine that some ML fairness systems also exhibit disparate treatment (see Section~\ref{sec:implicit_awareness} and Figure~\ref{fig:last_layer_tSNE});
and, more importantly, the decisions made by a fair regularized classifier are indistinguishable from those training a classifier designed to exhibit disparate treatment (see Section~\ref{sec:explicit_approach}). 

\subsection*{Summary} While it may not be possible to determine a priori if a particular fairness definition gives rise to disparate treatment, our decomposition of existing fair classifiers into an unconstrained classifier~$f$, and a second head that rescores the response using inferred race or gender, strongly aligns with the legal definition of disparate treatment. As such, we believe that this decomposition will be of value in determining the legality of deploying fair systems in practice. From a practical perspective, this disparate treatment is most strongly observed when the unconstrained classifier~$f$ has high demographic disparity, and demographic parity is strongly enforced. This makes it unlikely for any of the considered fairness methods to be appropriate tools to enforce equity without legal~reform.

\section{Other Related Work}\label{sec:related_work}

\textbf{Bias and Bias Amplification in Deep Neural Networks}
Numerous papers have found deep  networks discriminate based on protected groups \cite{klare2012face,gendershades,albiero2020analysis,feldman2021,balakrishnan2021towards} and even amplify bias present in the training data \cite{zhao2017men,hendricks2018,wang2019balanced,jia2020,wang2021,prates2020}.  Deep models have also been found to ``overlearn'', that is they learn representations encoding concepts that are not part of the learning objective; e.g., encoding race when trained to predict gender \cite{song2019overlearning,serna2020}. \cite{song2019overlearning} argue that overlearning is problematic from a privacy perspective as it reveals sensitive information. However, they do not consider if it allows models to disparately treat different groups. To detect unintended classifier bias, \cite{balakrishnan2021towards, denton2019image} synthesize counterfactual images by changing latent factors of a generative model, corresponding to attributes such as race, and seeing how performance alters. This is  related to our approach, however, they do not examine how fair models alter this behavior, and our decomposition of models into two heads allows us to reason counterfactually without generating images.

\textbf{Exploiting Disparate Treatment}
Several papers propose to use protected information for learning group-dependent models, in order to improve performance and / or fairness, assuming that doing so is legally acceptable and the 
protected information is available \cite{klare2012face,hardt2016equality,dwork18a,ustun19a}. 
Also the idea of using an estimate or score of the protected attribute is not new: \citet{Menon2018} and \citet{oneto2019} propose to infer the protected attribute from non-protected features and to use the inferred attribute to learn ``group specific'' models, as a way to enforce demographic parity or equalized odds. While these approaches are similar to our  approach presented in Section~\ref{sec:explicit_approach}, the  interpretation 
provided by \citeauthor{oneto2019}  is quite the opposite from ours: they consider their approach as a means of overcoming disparate treatment while we argue that such an approach should not be treated differently than an approach that explicitly uses protected~information.

\textbf{Bias Mitigation Methods} In the last few years, a plethora of fairness notions, that is definitions of fairness-concerning \emph{bias}, along with methods for mitigating such bias have been proposed, both in supervised and unsupervised learning. The methods in supervised learning are usually categorized into three groups: preprocessing methods, in-processing methods, and postprocessing methods (see \cite{caton2020,mehrabi2021} for survey papers). In this paper we study methods from each of the three groups (cf. Section~\ref{sec:preliminaries}): the regularizer approach belongs to the group of in-processing methods (and so does our strategy proposed in Section~\ref{sec:explicit_approach}), the massaging method of \cite{Kamiran2012} is a preprocessing method, and the strategy of \cite{lipton2018}  is a postprocessing method.   While the earlier papers on fair ML primarily considered tabular data, more recently, bias mitigation has also been studied in the context of deep learning \cite{du2020,wang2019balanced,wang2019racial,ramaswamy2020gandebiasing,wang2020towards}. \citet{fazelpour2020algorithmic} discuss a broader human-centered view going beyond altering algorithms with parity metrics.

\section{Implementation Details}\label{suppsec:technical_details}

Here we report details omitted, for reasons of space, from the body of the paper. In Section~\ref{suppsec:datasets_details} we give details about the datasets that were used.
In Section~\ref{suppsec:exp_details}, we give the formulation of a second regularizer, which is the absolute value of the relaxed DDP measure, and provide details on model training and the grid search procedure for our approach.

\subsection{Datasets}\label{suppsec:datasets_details}

\textbf{CelebA} The CelebA dataset \citep{liu2015faceattributes} contains $202,599$ images of celebrity faces with $40$ binary annotations, such as \textsc{Wearing\_glasses}, \textsc{Smiling} or \textsc{Male}. We use the Aligned\&Cropped subset and its standard split into train, test, and validation data. We center-crop the images and resize them to $224 \times 224$. During training, we randomly crop and flip images horizontally. We use \cite{ramaswamy2020gandebiasing} as reference for choosing target and protected labels.

\textbf{FairFace} The FairFace dataset~\citep{fairface}, published under CC BY 4.0 licence, contains $108501$ images  collected from the YFCC-100M Flickr dataset and are annotated with \textsc{gender}, \textsc{race}, and \textsc{age}. We binarize the attribute \textsc{race} into \textsc{White} and the union of all other groups. From \textsc{age}, we build several binary attributes: \textsc{Below\_20}, \textsc{Below\_30}, and \textsc{Below\_40}. In our experiments, we use the provided validation data with 1.25 padding as our test data, and from the provided train data, we prepared our own random and balanced validation split. We center-crop the images and resize them to $224 \times 224$. During training, we crop randomly, and flip the images horizontally with probability $0.5$.

\subsection{A Second Regularizer and Experimental Details.} \label{suppsec:exp_details}

In Section~\ref{sec:preliminaries} we have introduced the squared fairness regularizer~$\widehat{\mathcal{R}}_{\mathrm{DP}}$ (cf.~\eqref{def_fairness_regularizer}), used by \citet{wick2019}. We performed our experiments also with another regularizer \citep{padala2020fnnc}, denoted by $\widehat{\mathcal{R}}^{\mathrm{abs}}_{\mathrm{DP}}$. It is similar to  $\widehat{\mathcal{R}}_{\mathrm{DP}}$, but with the squaring function replaced by the absolute value, that is
\begin{align}\label{supp:def_fairness_regularizer_abs}
     \widehat{\mathcal{R}}^{\mathrm{abs}}_{\mathrm{DP}}(f) := \left|\frac{1}{|\{s: s_i=1\}|} \sum_{i \in [n]: s_i = 1} \sigma(f(x_i))- \frac{1}{|\{s: s_i=0\}|} \sum_{i \in [n]: s_i = 0} \sigma(f(x_i))\right|.
\end{align}

\textbf{Models and Optimization} Given a fixed target attribute and protected attribute, we train all parameters of a  pretrained ResNet50~\citep{He2016resnet} or MobileNetV3-Small~\citep{howard2019mobilenet} backbone provided by PyTorch with binary cross entropy loss. MobileNetV3-Small contains $2.8$M parameters and is more resource friendly than the much bigger ResNet50. Hence, for some experiments we only used MobileNetV3-Small to save computation time. The dimension $m$ of the last-layer representation $z\in \R^m$ is $m=2048$ for the ResNet50 and $m=1024$ for MobileNetV3-Small.

We train all models, including our approach, with the Adam Optimizer \citep{kingma2017adam} (learning rate is $10^{-4}$ on CelebA and $10^{-5}$ on FairFace, batchsize is $64$) for a total of $20$ epochs and select the model with the highest average precision achieved on the validation set. In addition, we employ a learning rate scheduler that reduces the learning rate by a factor of $10$ if there is no progress on the validation loss for more than $8$ epochs. To have meaningful regularizer losses for each batch, we use stratified batches, such that the prevalence of the protected attribute is roughly the same as the overall prevalence. For the classification loss, we use binary cross entropy loss with a sigmoid activation. 

If we train the models with one of our two fairness regularizers, the range for the fairness parameter~$\lambda$ is $\lambda \in [0, 0.1, 0.5, 1, 2, 3, 4, 5, 10, 15, 20, 30]$. For the \emph{Massaging} preprocessing method, the range is $\lambda \in[0, 0.1, 0.2, 0.3, 0.4, 0.5, 0.6, 0.7, 0.8, 0.9, 1.0]$.

\textbf{Grid Search} The grid search procedure of our  approach chooses all combinations of $a_1$ and $a_2$ from a grid of $200$ equidistant points between $-15$ and $15$. Going through all combinations, we choose the most accurate model that satisfies the user-chosen fairness constraint. We continue to search in the interval of the grid points which are closest to the current solution by forming another grid of $200$ equidistant points in this interval. We continue this recursion $4$ times.

\FloatBarrier

\section{Extended Experimental Results}\label{suppsec:extended_results}

In Section~\ref{suppsec:implicit_awareness}, we complement the results on protected attribute awareness in fair networks including results using the second regularizer \eqref{supp:def_fairness_regularizer_abs}. In Section~\ref{suppsec:explicit_model}, we apply our explicit approach to MobileNetV3-Small and compare to all other fairness approaches. Finally, in Section~\ref{suppsec:model_equivalence}~and~\ref{suppsec:disparate_treatment} we extend our main result about identifying disparate treatment to different target tasks, models, and fairness approaches.

\subsection{Protected Attribute Awareness}\label{suppsec:implicit_awareness}

\begin{figure}[h]
\centering
\includegraphics[width=.9\textwidth]{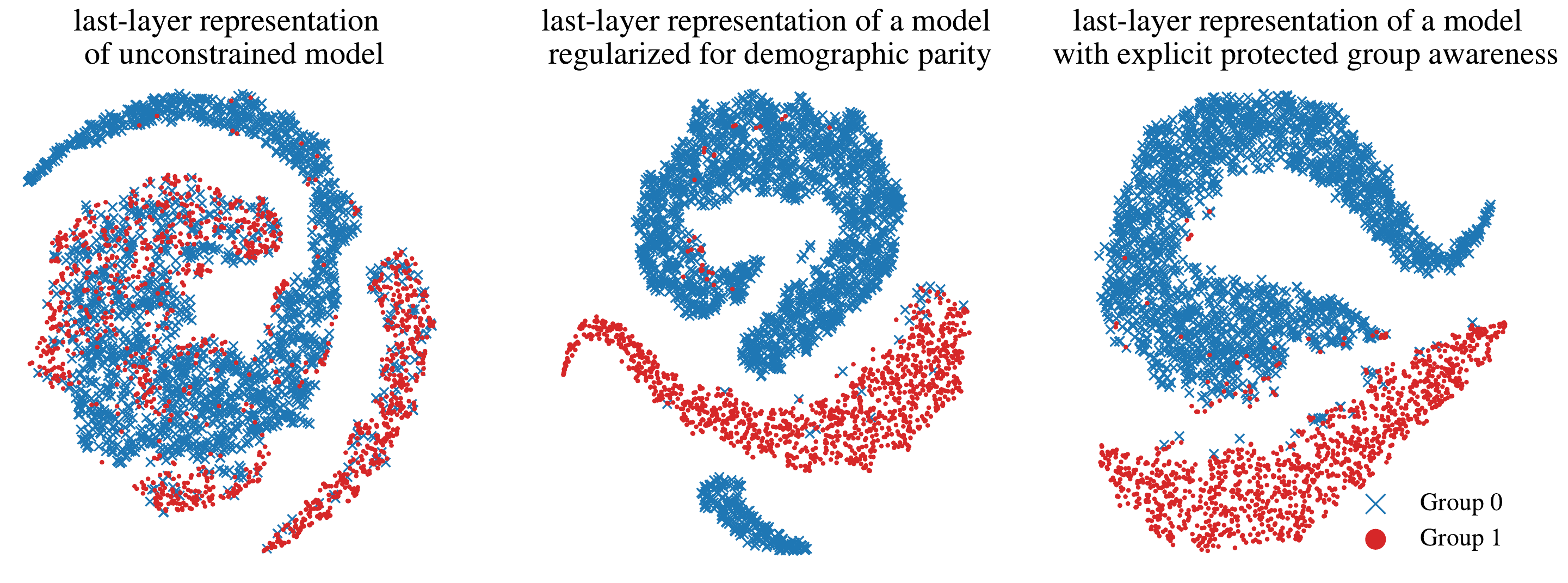}
\caption{{\bf tSNE~\cite{maaten2008tsne} visualization of feature representations of unconstrained (left), fairness-regularized with $\widehat{\mathcal{R}}_{\mathrm{DP}}$ (center), and group-aware (Section~\ref{sec:explicit_approach}) (right) Resnet50 models.} Each point is colored according to the protected attribute \textsc{Male}, and we aim to classify the binary label \textsc{Attractive}. Similar to the main paper, we observe that the fair model in the center and the group aware model on the right separate the genders.} \label{suppfig:last_layer_tSNE_attractive}
\end{figure}

In this section, we report further results on protected attribute awareness in fair neural networks. We plot last-layer tSNE visualizations for another CelebA task in Figure~\ref{suppfig:last_layer_tSNE_attractive}~and for a FairFace task in Figure~\ref{suppfig:last_layer_tSNE_attractive}. Similar to Figure~\ref{fig:last_layer_tSNE} in the body of the paper, gender is separated into two clusters when we regularize the model for demographic parity. 

\begin{figure}[hb!]
\centering
\subcaptionbox{MobileNetV3-Small with regularizer $\widehat{\mathcal{R}}_{\mathrm{DP}}$ on CelebA. }{\includegraphics[height=7cm]{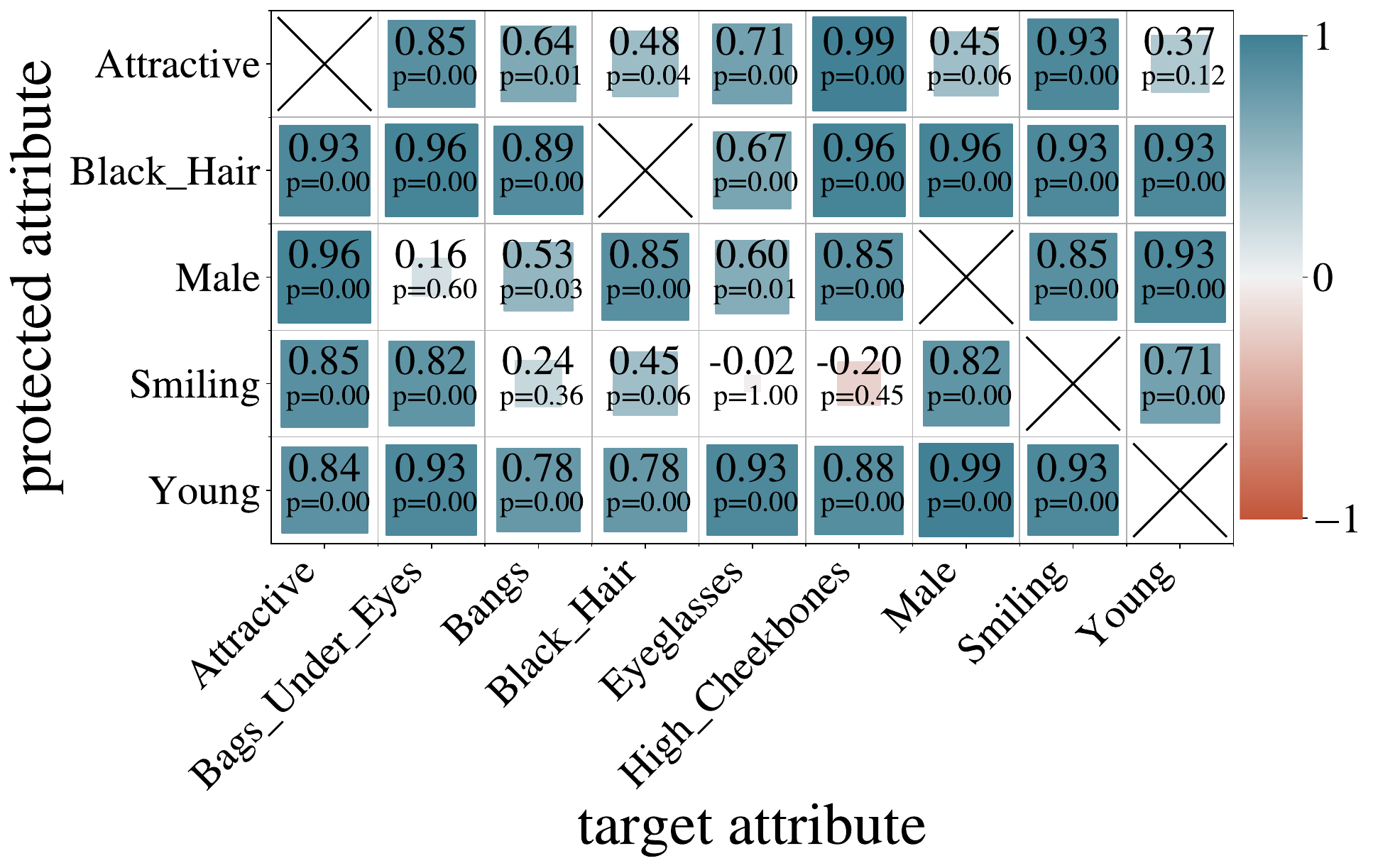}}\vspace{1cm}
\subcaptionbox{ MobileNetV3-Small with regularizer $\widehat{\mathcal{R}}^{\mathrm{abs}}_{\mathrm{DP}}$ on CelebA.}{\includegraphics[height=7cm]{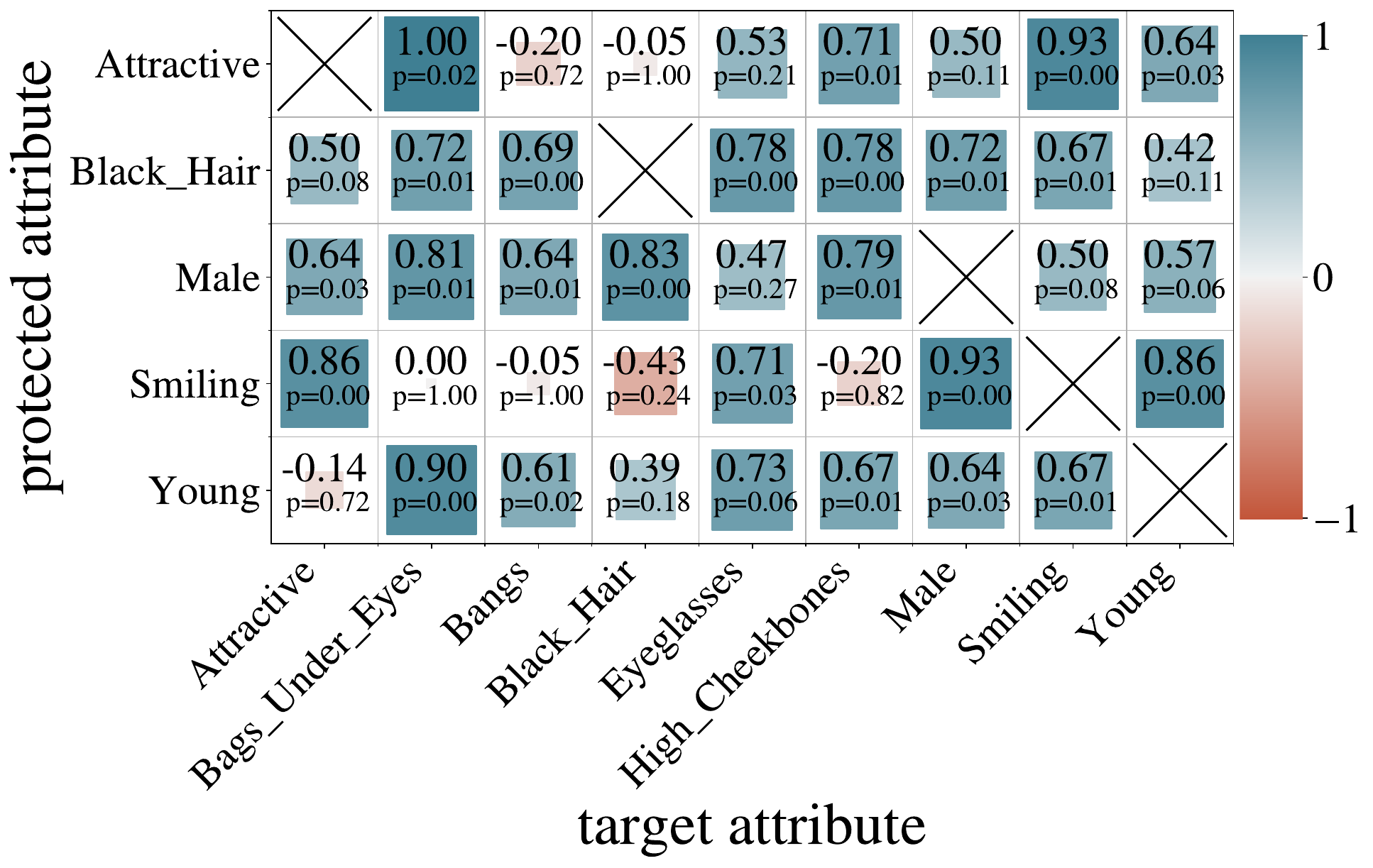}}
    \caption{\textbf{Kendall-tau correlation between fairness parameter and protected attribute accuracy.} Similar to the results in the main paper, where ResNet50 was used, we also find for MobileNetV3-Small that group awareness is increasing as the fairness parameter is increased. In (b) we evaluate the regularizer $\widehat{\mathcal{R}}^{\mathrm{abs}}_{\mathrm{DP}}$ and, although on fewer tasks, observe a similar behavior.}
    \label{suppfig:correlation_table_mobilenet}
\end{figure}

\begin{figure}[hb!]
\centering
\subcaptionbox{MobileNetV3-Small with \emph{Massaging} preprocessing on CelebA.}{\includegraphics[width=0.6\textwidth]{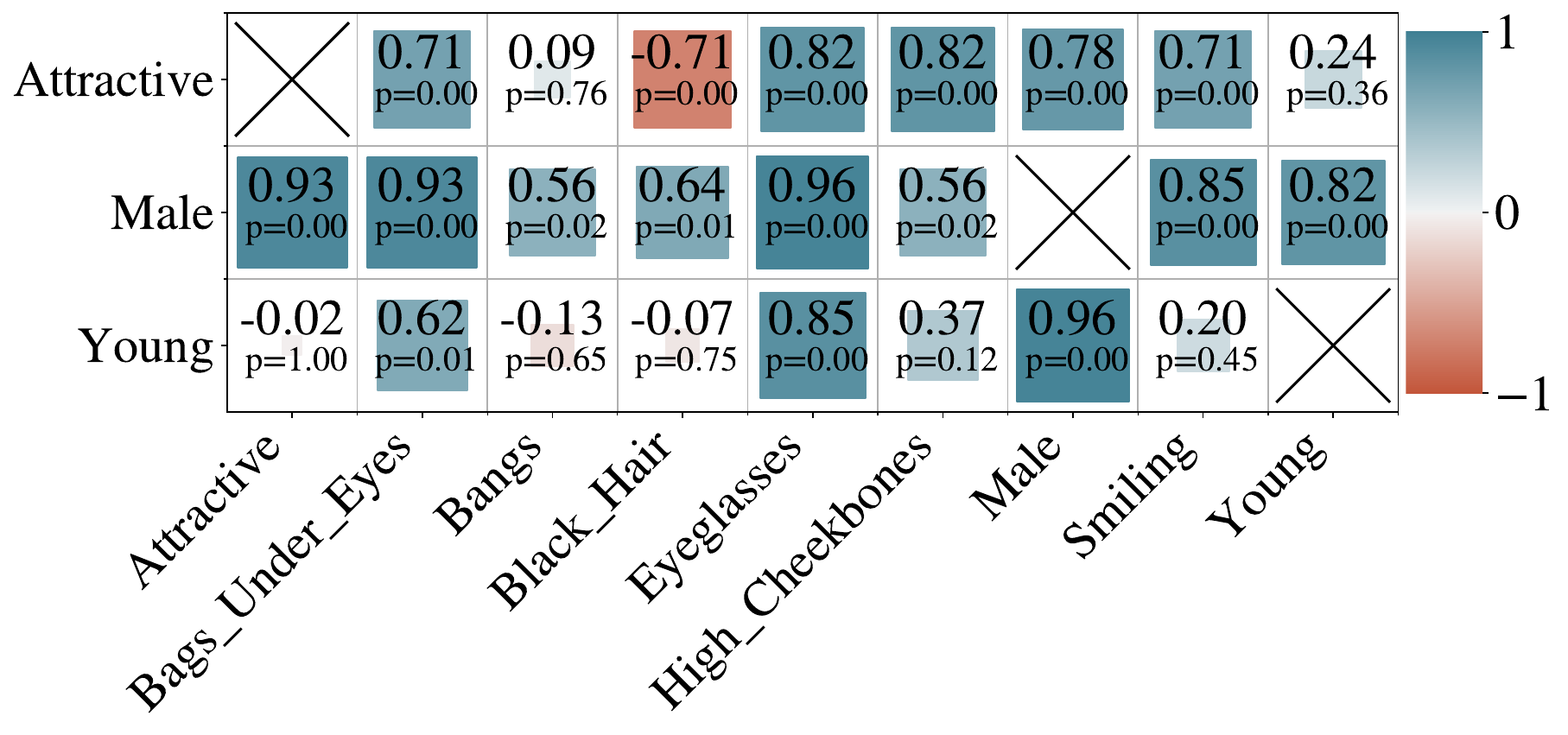}}
    \caption{\textbf{Kendall-tau correlation between fairness parameter $\lambda$ and protected attribute accuracy.} Similar to the regularized approaches, we find an increased group awareness for the \emph{Massaging} preprocessing method, especially when the protected attribute is \textsc{Male}.}
    \label{suppfig:correlation_table_massaging}
\end{figure}

\begin{figure}
\centering
\subcaptionbox{MobileNetV3-Small with regularizer $\widehat{\mathcal{R}}_{\mathrm{DP}}$ on FairFace. }{\includegraphics[width=0.4\textwidth]{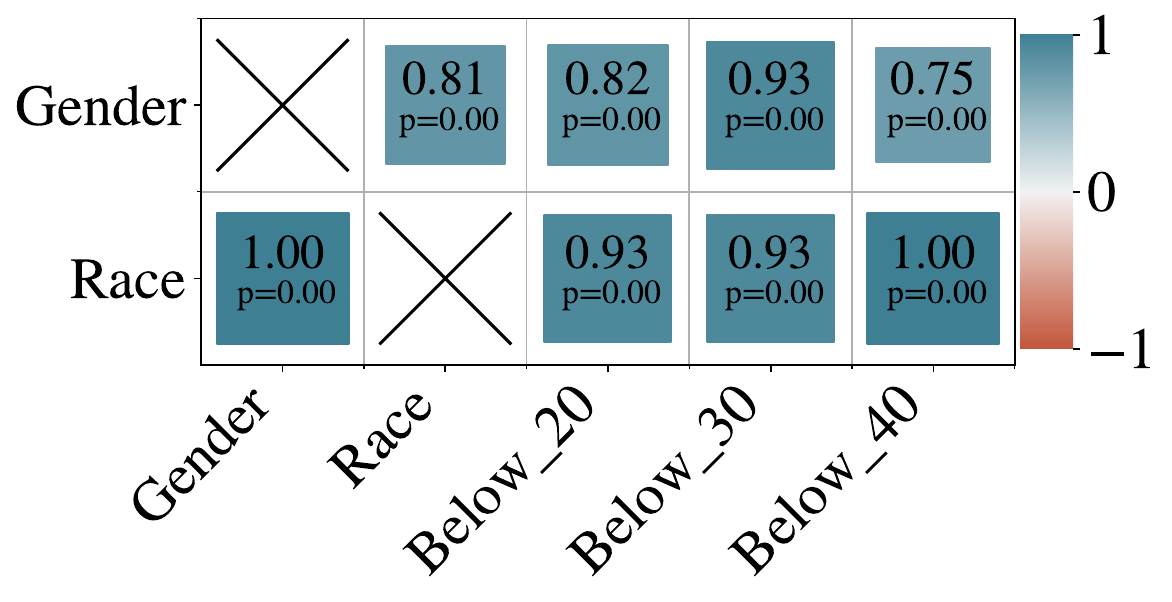}}\hspace{1cm}
\subcaptionbox{MobileNetV3-Small with regularizer $\widehat{\mathcal{R}}^{\mathrm{abs}}_{\mathrm{DP}}$ on FairFace.}{\includegraphics[width=0.4\textwidth]{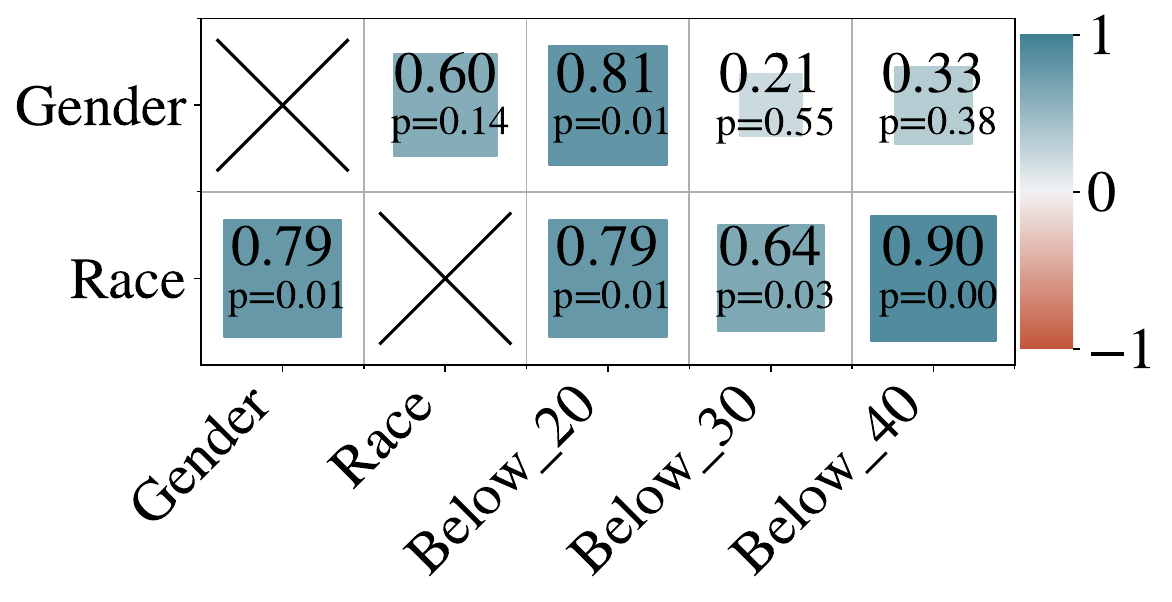}}
    \caption{\textbf{Kendall-tau correlation between fairness parameter $\lambda$ and protected attribute accuracy.} Similar to the findings on the CelebA dataset, we also find an increased group awareness on FairFace for the protected attributes \textsc{Race} and \textsc{Gender}.}
    \label{suppfig:correlation_table_mobilenet_fairface}
\end{figure}

\begin{figure}
\centering
\subcaptionbox{ResNet50 with regularizer $\widehat{\mathcal{R}}_{\mathrm{DP}}$ on FairFace.}{\includegraphics[width=0.4\textwidth]{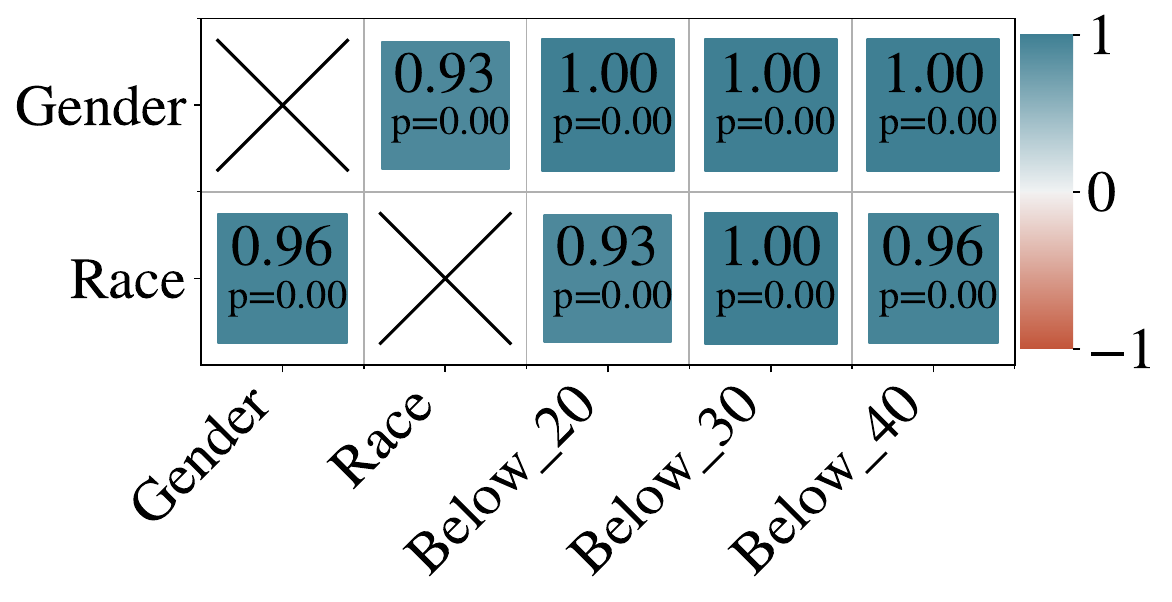}}\hspace{1cm}
\subcaptionbox{ResNet50 with regularizer $\widehat{\mathcal{R}}_{\mathrm{DP}}$ on FairFace.}{\includegraphics[width=0.3\textwidth]{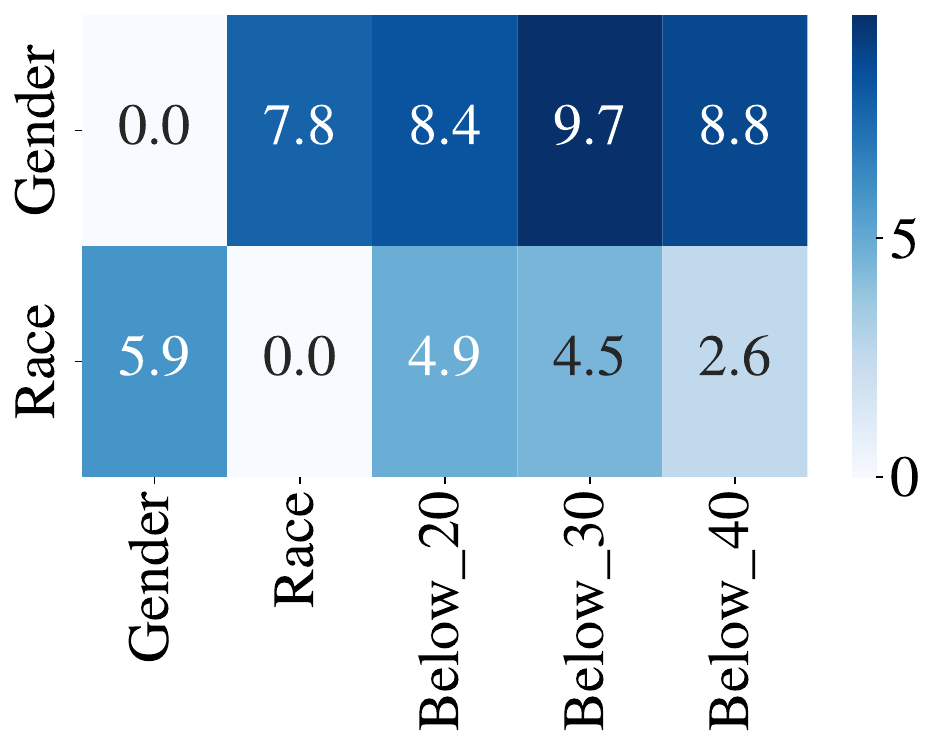}}
    \caption{(Left) \textbf{Kendall-tau correlation between fairness parameter $\lambda$ and protected attribute accuracy.} (Right) \textbf{Increase of protected attribute accuracy} of the group classifier learned on the last layer of ResNet50.}
    \label{suppfig:correlation_table_resnet_fairface}
\end{figure}

In Figure~\ref{suppfig:correlation_table_mobilenet}, we plot the Kendall-tau correlations when using MobileNetV3-Small for both of the two presented regularizers. As with a ResNet50 model, we observe a strong association between fairness parameter and an increase in group awareness. However, for the $\widehat{\mathcal{R}}^{\mathrm{abs}}_{\mathrm{DP}}$ regularizer a positive association is less often significant than for $\widehat{\mathcal{R}}_{\mathrm{DP}}$. In Figure~\ref{suppfig:correlation_table_massaging}, we apply \emph{Massaging} preprocessing with a varying fairness parameter. Results on FairFace are presented in Figures~\ref{suppfig:correlation_table_mobilenet_fairface} and ~\ref{suppfig:correlation_table_resnet_fairface}.

\newpage

\begin{figure}[h]
\centering
\includegraphics[width=.9\textwidth]{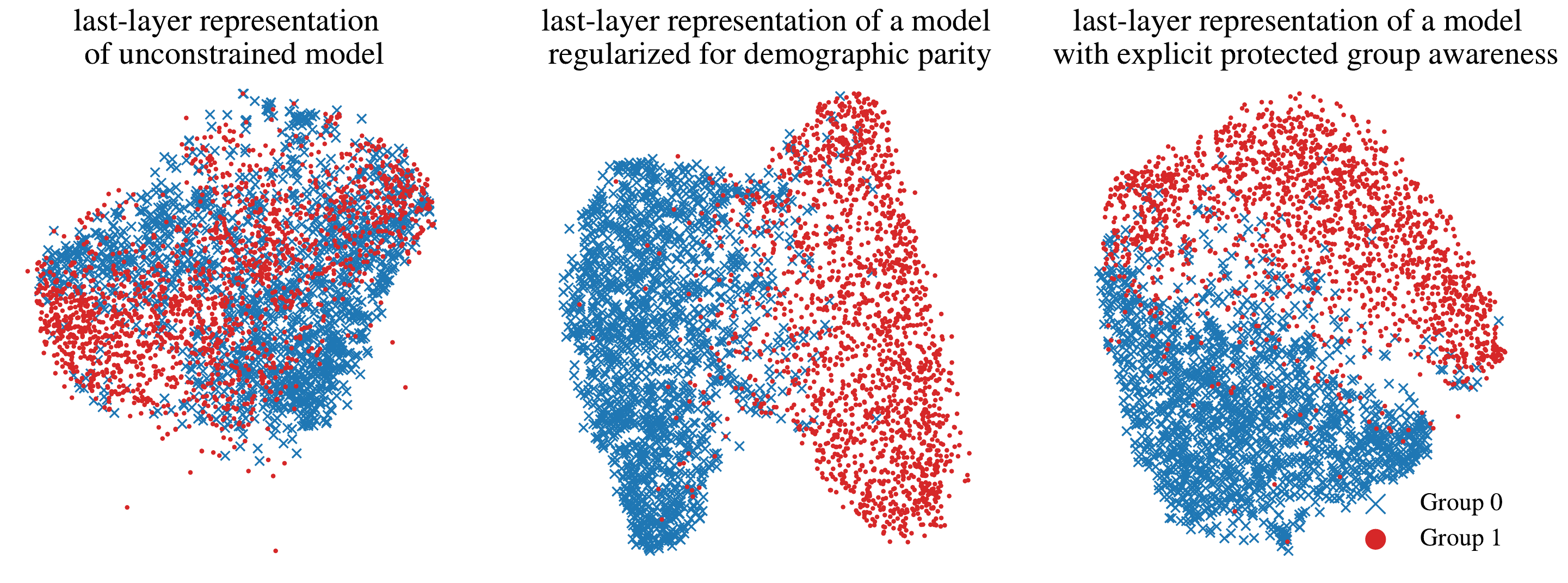}
\caption{{\bf tSNE visualization of feature representations of unconstrained (left), fairness-regularized (center), and group-aware (right) Resnet50 models.} In this figure, we use the FairFace dataset. Each point is colored according to the protected attribute \textsc{Gender}, and we classify the binary label \textsc{Below\_30}. Similar to CelebA, we observe that gender is separated into disjoint clusters in fair and group aware models, whereas they were mixed in the unconstrained model.} \label{suppfig:last_layer_tSNE_fairface}
\end{figure}

\begin{figure}
    \centering
    \subcaptionbox{Demographic parity violation (DDP) of unconstrained ResNet50 on CelebA.}{\includegraphics[height=7cm]{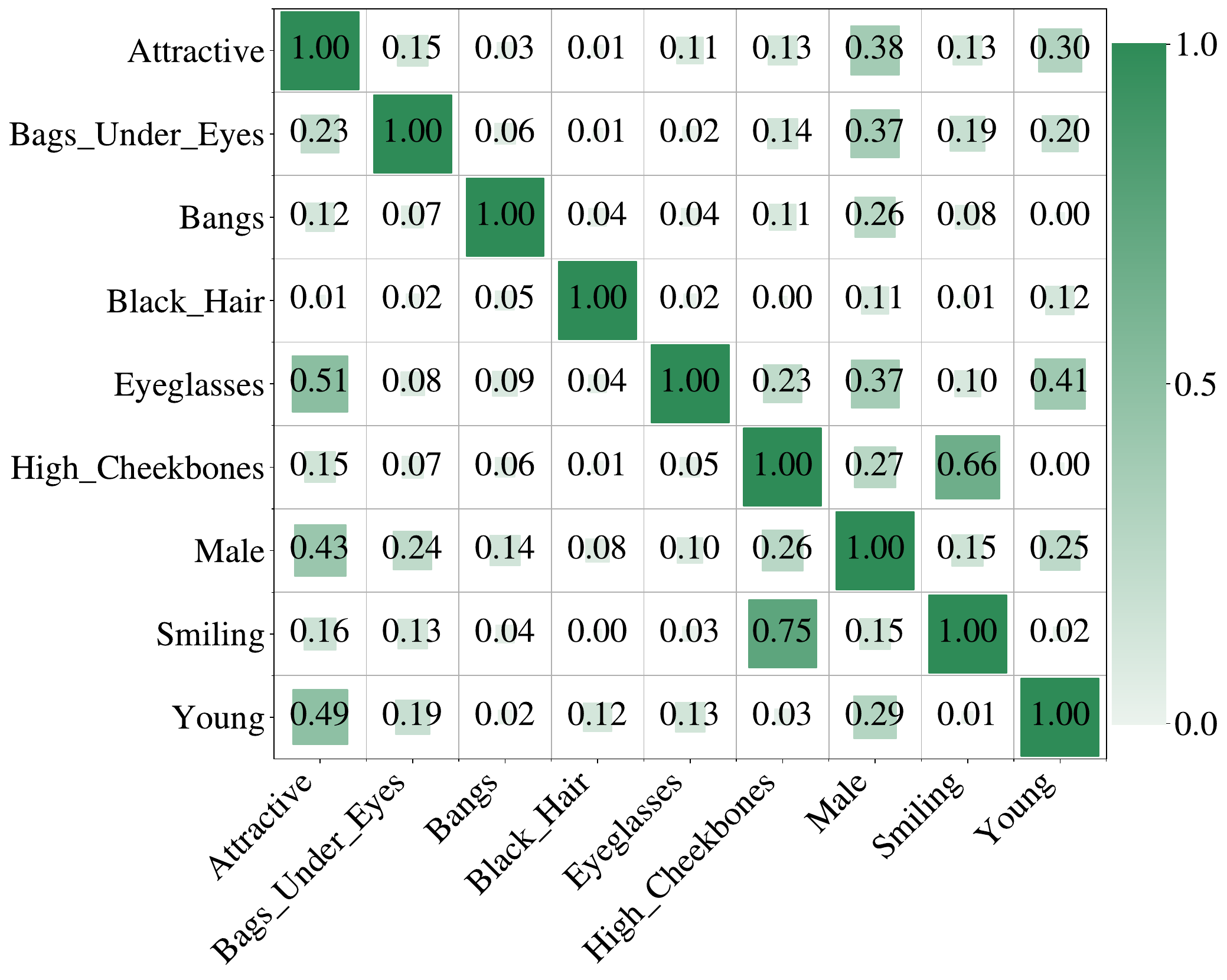}}\vspace{0.5cm}
    \subcaptionbox{Maximum increase of protected attribute accuracy when training ResNet50 with regularizer $\widehat{\mathcal{R}}_{\mathrm{DP}}$ on CelebA.}{\includegraphics[height=7.5cm]{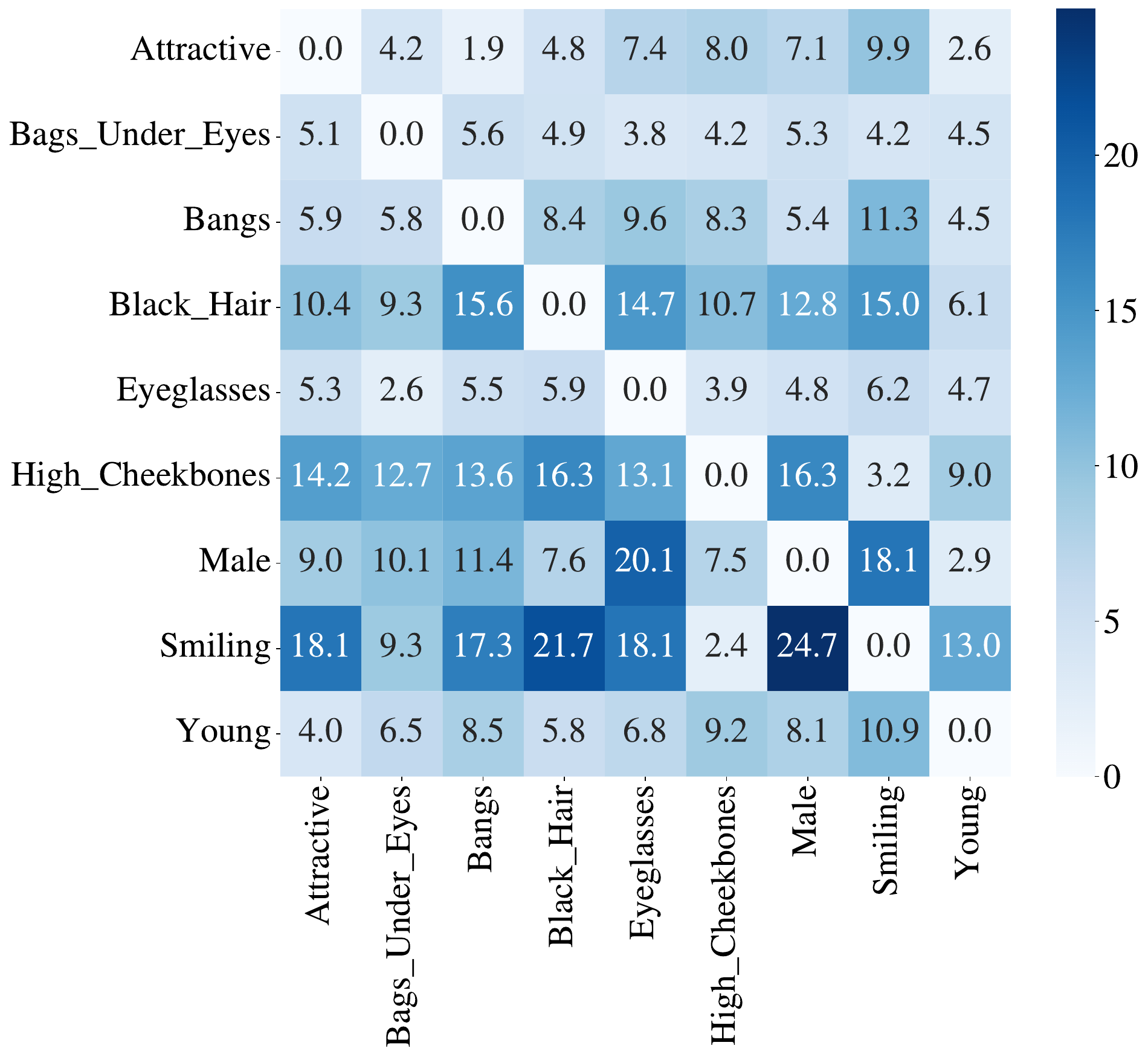}}
    \caption{ (Top) \textbf{Demographic parity violation (DDP) of the unconstrained classifier.} The increase in group awareness is more moderate for those tasks where the unconstrained classifier is very unfair, for example for the task (column) \textsc{Male}. However, this is not always the case as for protected attribute \textsc{Smiling} and target \textsc{Bangs} for example. (Bottom) \textbf{Maximum increase of protected attribute accuracy.} Compared to the unconstrained model, we show the highest difference to the second head accuracy of fair models. Even though the unconstrained model is fair, for example for a few target tasks when protected attribute is \textsc{Smiling}, the increase in second head accuracy can still be large.}
    \label{suppfig:accuracy_increase}
\end{figure}

\newpage
\FloatBarrier

\subsection{Our Explicit Approach.}\label{suppsec:explicit_model}

\def\rot{\rotatebox}

 \begin{table}[ht]
        \caption{\textbf{Accuracy under strict fairness constraints.} The first block requires a reduction of the absolute value of the DDP of at least $50$\%, the second at least $80$\%. The protected attribute is \textsc{Male} from CelebA dataset and we use the MobileNetV3-Small architecture. Crosses indicate that the method did not achieve the required fairness. Similar to the main paper, the regularizer $\widehat{\mathcal{R}}_{\mathrm{DP}}$ often fails to find sufficiently fair solution. The regularizer $\widehat{\mathcal{R}}^{\mathrm{abs}}_{\mathrm{DP}}$ always finds fair solutions, however, at high costs in accuracy, often resulting in trivial solutions. \textbf{Our explicit two-headed approach can always find a fair solution} and is comparable to Lipton, which, contrary to us, requires the true protected attribute.}
        \centering
        \resizebox{\columnwidth}{!}{
        \begin{tabular}{@{}c c@{}c@{}c@{}c@{}c@{}c@{}c@{}c@{}}&\rot{35}{Attractive}&\rot{35}{Bags\_Under\_Eyes}&\rot{35}{Bangs}&\rot{35}{Black\_Hair}&\rot{35}{Eyeglasses}&\rot{35}{High\_Cheekbones}&\rot{35}{Smiling}&\rot{35}{Young}\\\toprule
        &&&\multicolumn{3}{c}{50\% disparity reduction} \\\midrule
        Lipton& 0.8034& 0.8441& 0.9473& 0.9001& 0.9658& 0.8609& 0.9200& 0.8773\\Our Approach& 0.8023& 0.8439& 0.9445& 0.8930& 0.9773& 0.8613& 0.9212& 0.8762\\Massaging& 0.7986& 0.8424& 0.9396& 0.9012& 0.9661& 0.8572& 0.9135& 0.8520\\Regularizer $\widehat{\mathcal{R}}_{\mathrm{DP}}$& 0.7959& 0.8446& $\Cross$& 0.8993& $\Cross$& 0.8603& $\Cross$& 0.8710\\Regularizer $\widehat{\mathcal{R}}^{\mathrm{abs}}_{\mathrm{DP}}$& 0.7935& 0.8311& 0.8443& 0.8962& 0.9545& 0.8609& 0.4997& 0.8728\\ \midrule
&&&\multicolumn{3}{c}{80\% disparity reduction}\\\midrule
            Lipton& 0.7775& 0.8352& 0.9331& 0.9001& 0.9618& 0.8403& 0.9106& 0.8606\\Our Approach& 0.7741& 0.8347& 0.9310& 0.8921& 0.9652& 0.8443& 0.9105& 0.8580\\Massaging& 0.7612& 0.8204& $\Cross$& 0.8947& $\Cross$& $\Cross$& $\Cross$& 0.8468\\Regularizer $\widehat{\mathcal{R}}_{\mathrm{DP}}$& 0.7740& 0.8294& $\Cross$& 0.8989& $\Cross$& $\Cross$& $\Cross$& 0.8562\\ Regularizer $\widehat{\mathcal{R}}^{\mathrm{abs}}_{\mathrm{DP}}$& 0.7693& 0.8311& 0.8443& 0.8962& 0.9407& 0.5182& 0.4997& 0.8307\\
        \end{tabular}
        }
        \label{tab:two_heads_comparison_mobilenet}  
    \end{table}

\begin{figure}
    \centering
    \includegraphics[width=\textwidth]{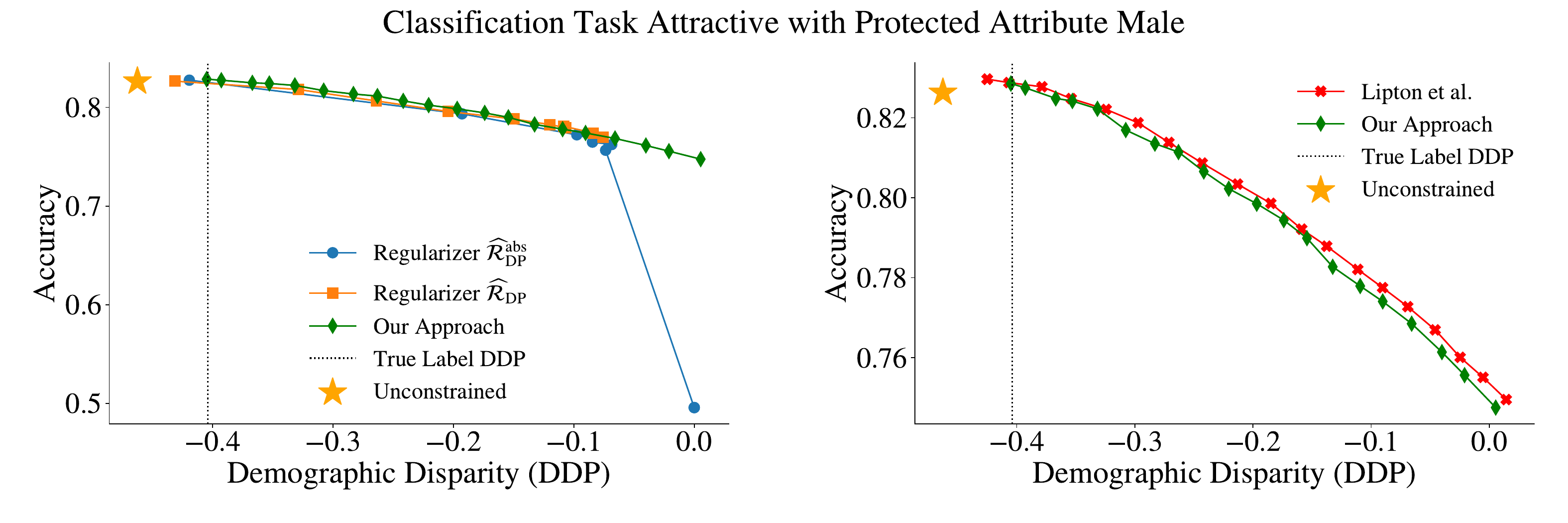}
    \caption{\textbf{Comparison of different fairness approaches using the MobileNetV3-Small architecture.} 
    We compare our group aware model to fairness-regularized models (\emph{left plot}) 
    and the approach of \citet{lipton2018} (\emph{right plot}) 
    on when predicting the target \textsc{Attractive} with respect to the protected attribute  \textsc{Male}.
    For all methods, we observe the typical trade-off:  as the model becomes fairer (DDP is closer to 0), the target accuracy for \textsc{Attractive} decreases. All methods obtain  similar accuracy for a particular DDP value. However, the regularizer $\widehat{\mathcal{R}}_{\mathrm{DP}}$ is unable to achieve near perfect fairness and saturates around a DDP value of~$-0.1$. The regularizer $\widehat{\mathcal{R}}^{\mathrm{abs}}_{\mathrm{DP}}$ collapses to a trivial fair solution. Note that \citet{lipton2018} 
    requires the protected attribute at test time, while we infer the protected attribute.}
    \label{suppfig:fairness_vs_accuracy_mobilenet}
\end{figure}
 
\newpage
\FloatBarrier

\subsection{Fair Networks Behave like our Explicit Approach.}\label{suppsec:model_equivalence}

\begin{figure}[h!]
    \centering
    \subcaptionbox*{}{\includegraphics[width=0.4\textwidth]{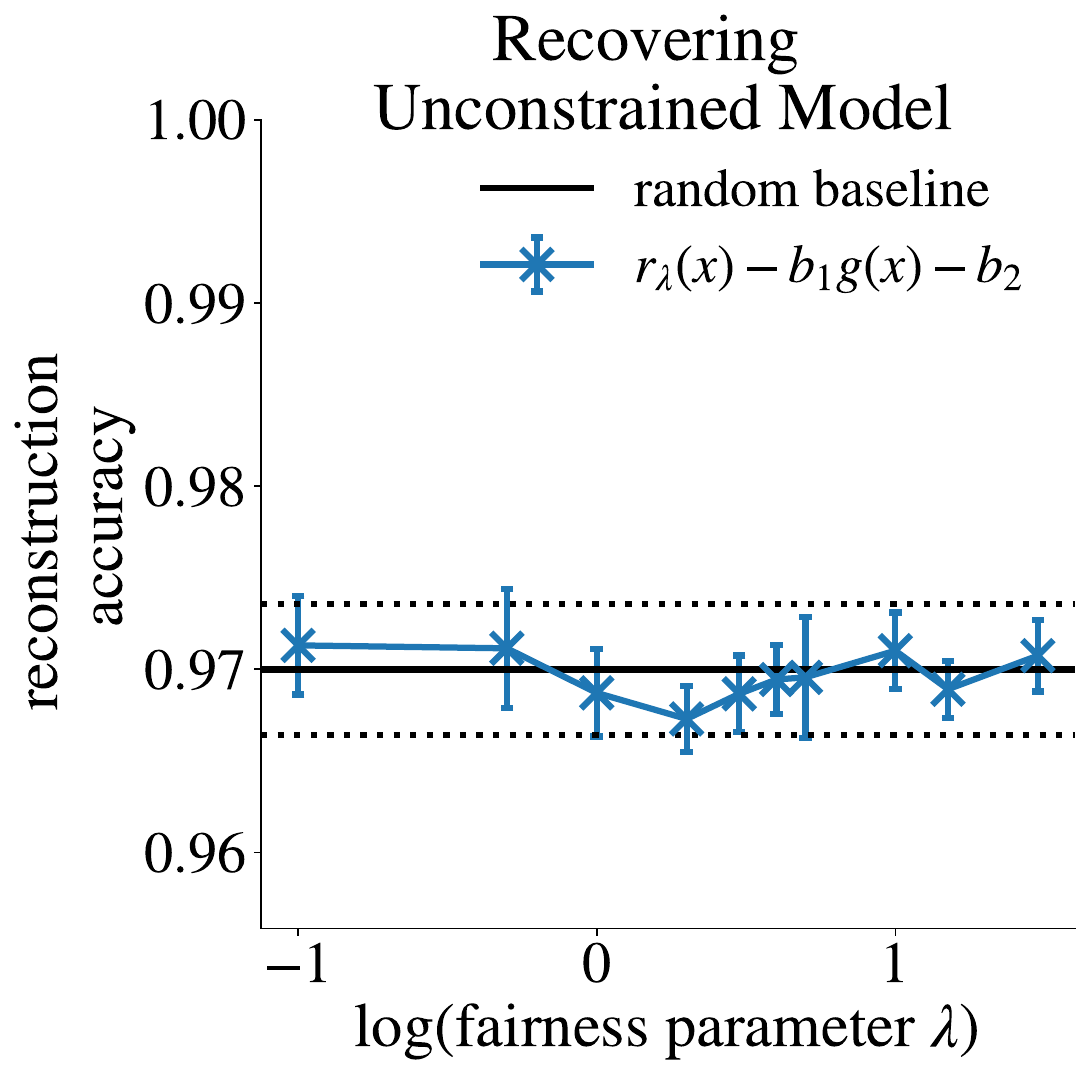}}\hspace{1cm}
    \subcaptionbox*{}{    \includegraphics[width=0.4\textwidth]{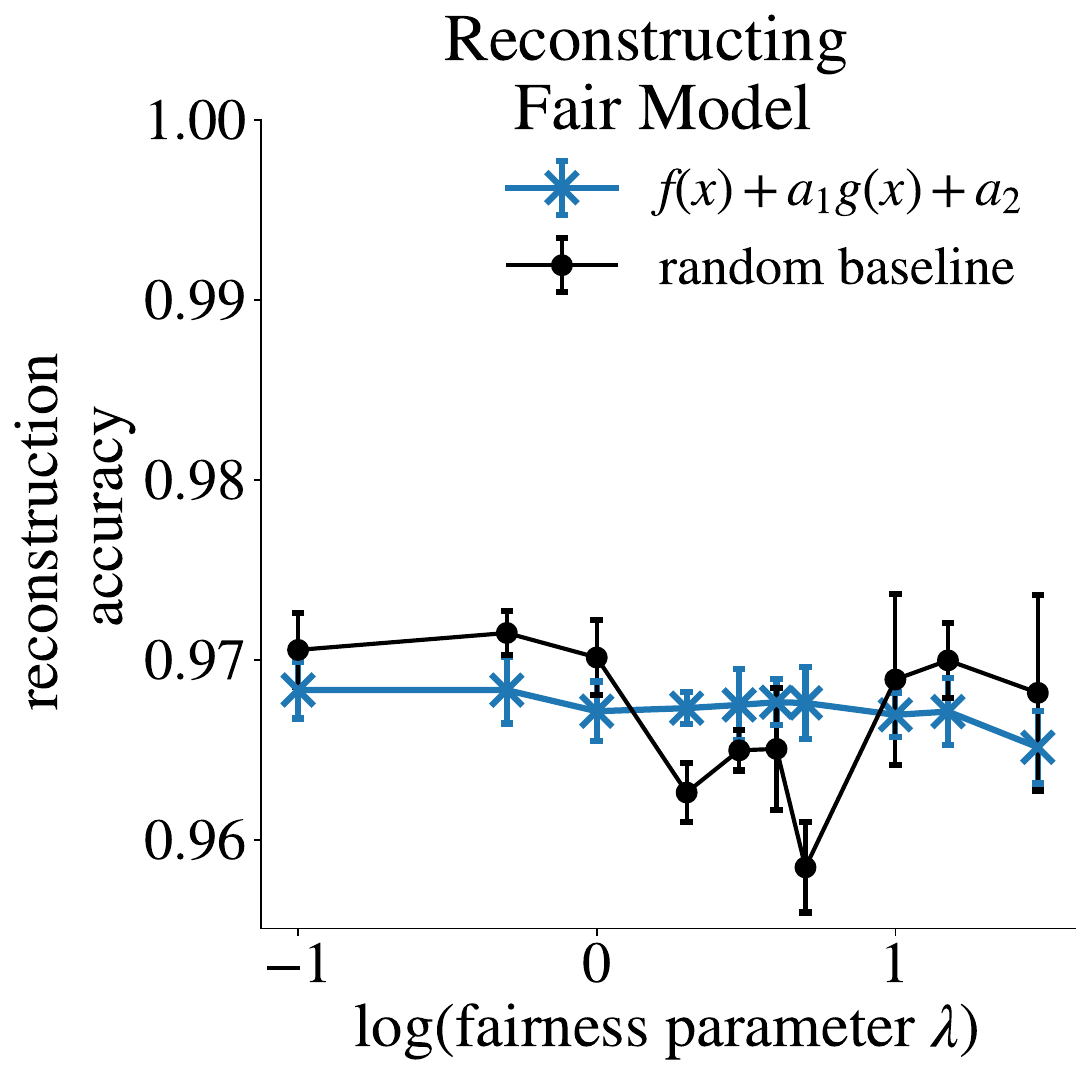}}
    \caption{\textbf{Recovering the unconstrained classifier and reconstructing fair classifiers.} We train ResNet50 models with the $\widehat{\mathcal{R}}_{\mathrm{DP}}$ regularizer for the target \textsc{Smiling} and protected attribute \textsc{Male}. Again, we can reconstruct fair models with our  approach and we can recover the unconstrained model by adding the second head to the fair model.
    }\label{suppfig:model_equivalence_smiling_resnet}
\end{figure}

\begin{figure}[h!]
    \centering
    \subcaptionbox*{}{\includegraphics[width=0.4\textwidth]{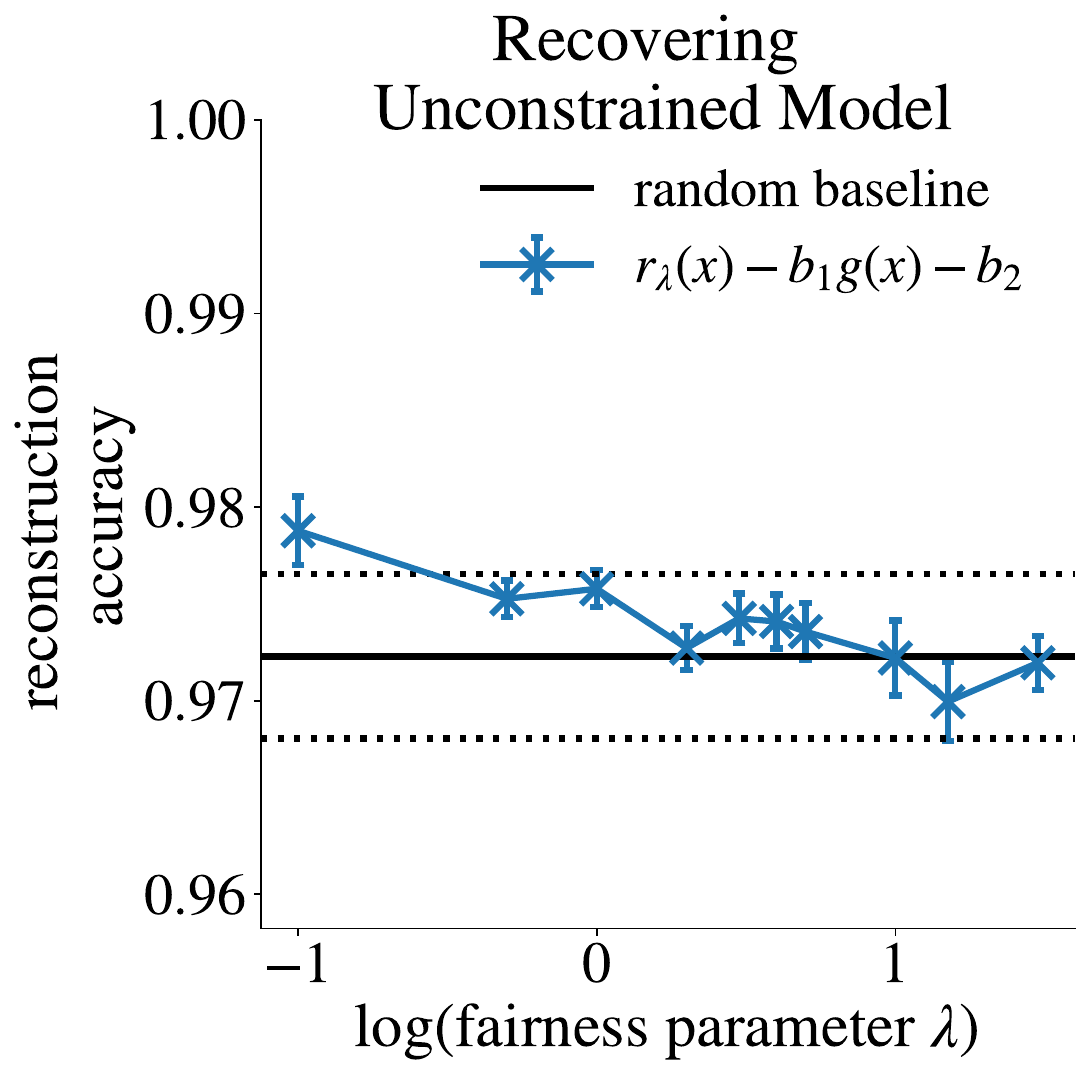}}\hspace{1cm}
    \subcaptionbox*{}{    \includegraphics[width=0.4\textwidth]{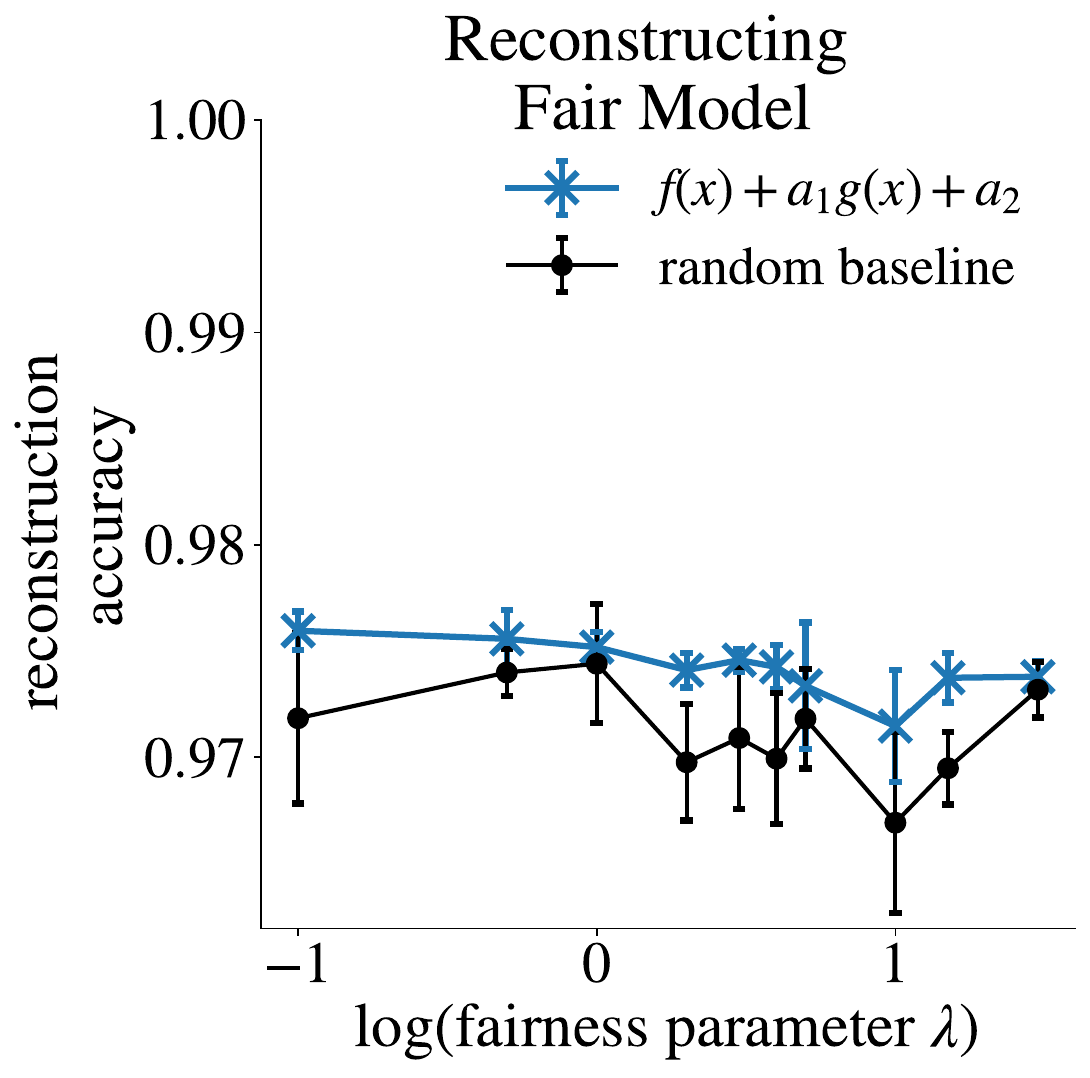}}
    \caption{\textbf{Recovering the unconstrained classifier and reconstructing fair classifiers.} We train MobileNetV3-Small models with the $\widehat{\mathcal{R}}_{\mathrm{DP}}$ regularizer for the target \textsc{Smiling} and protected attribute \textsc{Male}.  Similarly to the analysis above with a ResNet50, our observations hold for MobileNetV3-Small models as well.
    }\label{suppfig:model_equivalence_smiling_mobile}
\end{figure}

\begin{figure}
    \centering
    \subcaptionbox*{ }{\includegraphics[width=0.3\textwidth]{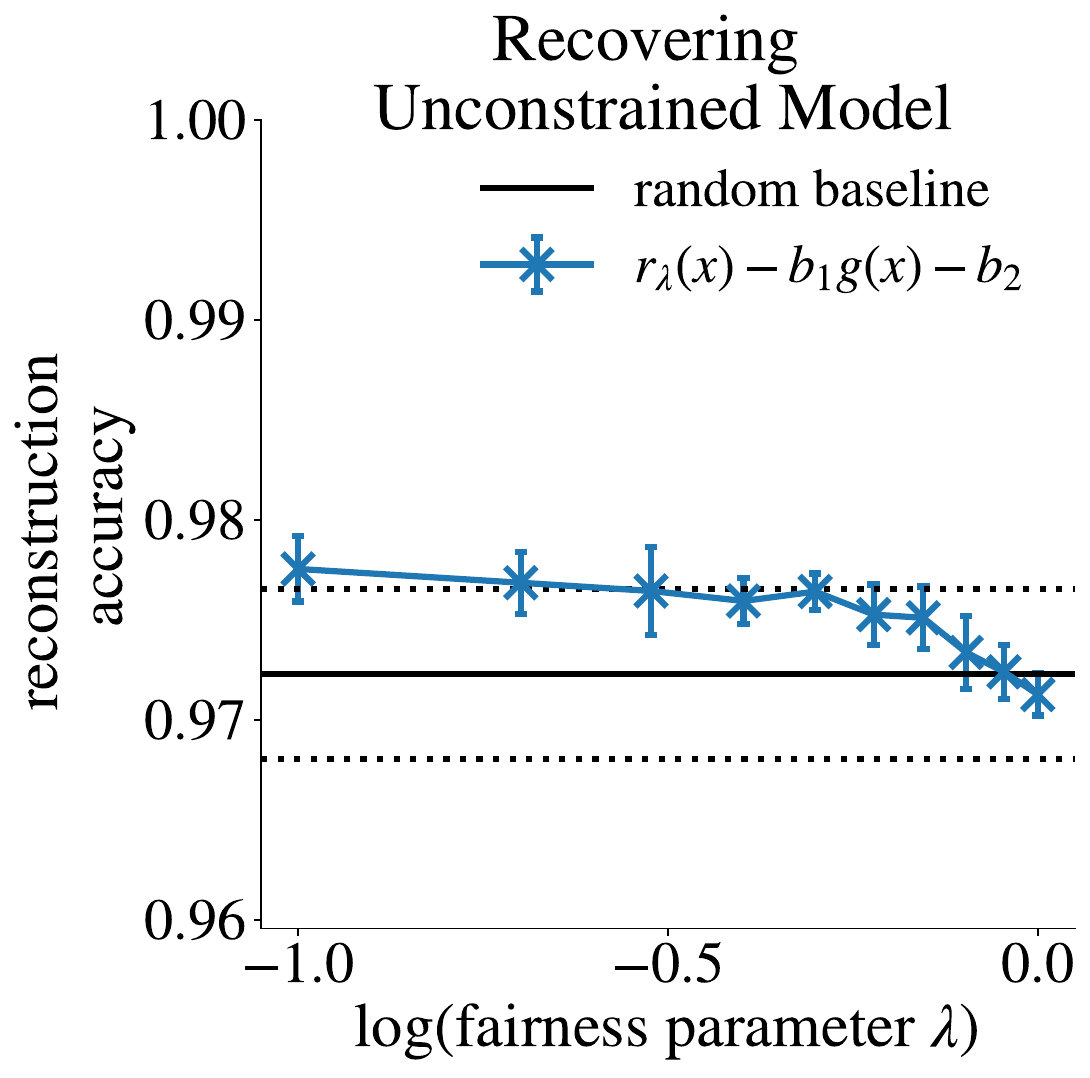}}\hspace{1cm}
    \subcaptionbox*{}{    \includegraphics[width=0.3\textwidth]{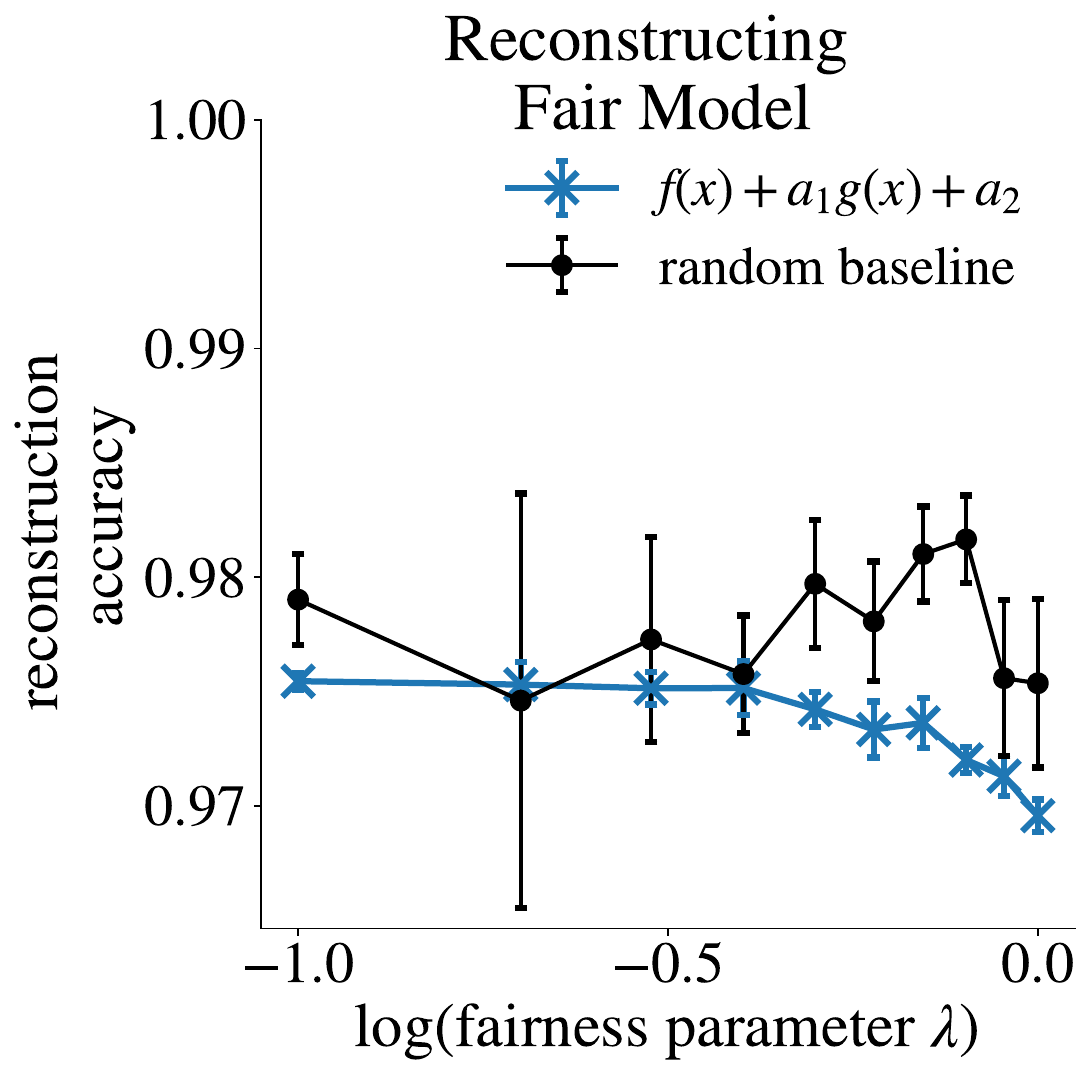}}
    \caption{\textbf{Recovering the unconstrained classifier and reconstructing fair classifiers.} We train MobileNetV3-Small models with the \emph{Massaging} preprocessing for the target \textsc{Smiling} and protected attribute \textsc{Male}. When using \emph{Massaging}, we can reconstruct the resulting fair models with our approach. However, it reconstructs the most fair models slightly less accurately than a retrained fair model.
    }\label{suppfig:model_equivalence_smiling_mobile_massaging}
\end{figure}

\begin{figure}
    \centering
    \subcaptionbox{ResNet50 with $\widehat{\mathcal{R}}_{\mathrm{DP}}$. }{\includegraphics[width=0.3\textwidth]{plots/model_equivalence/Male_Attractive/reconstr_acc_curve_resnet50_test_DDP_matkle_v1.pdf}}
    \subcaptionbox{MobileNetV3-Small with $\widehat{\mathcal{R}}_{\mathrm{DP}}$.}{    \includegraphics[width=0.3\textwidth]{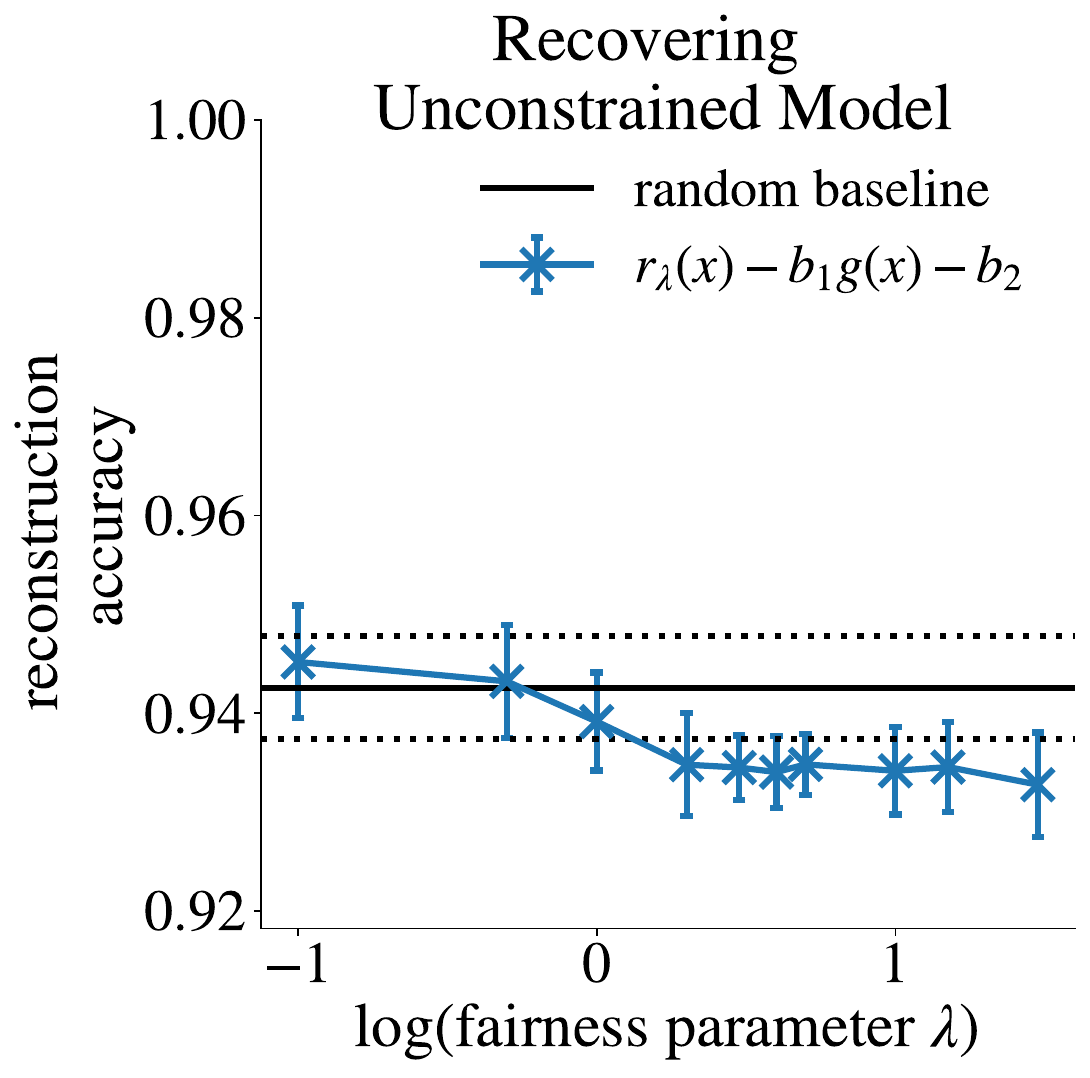}}
    \subcaptionbox{MobileNetV3-Small with \emph{Massaging}.}{
    \includegraphics[width=0.3\textwidth]{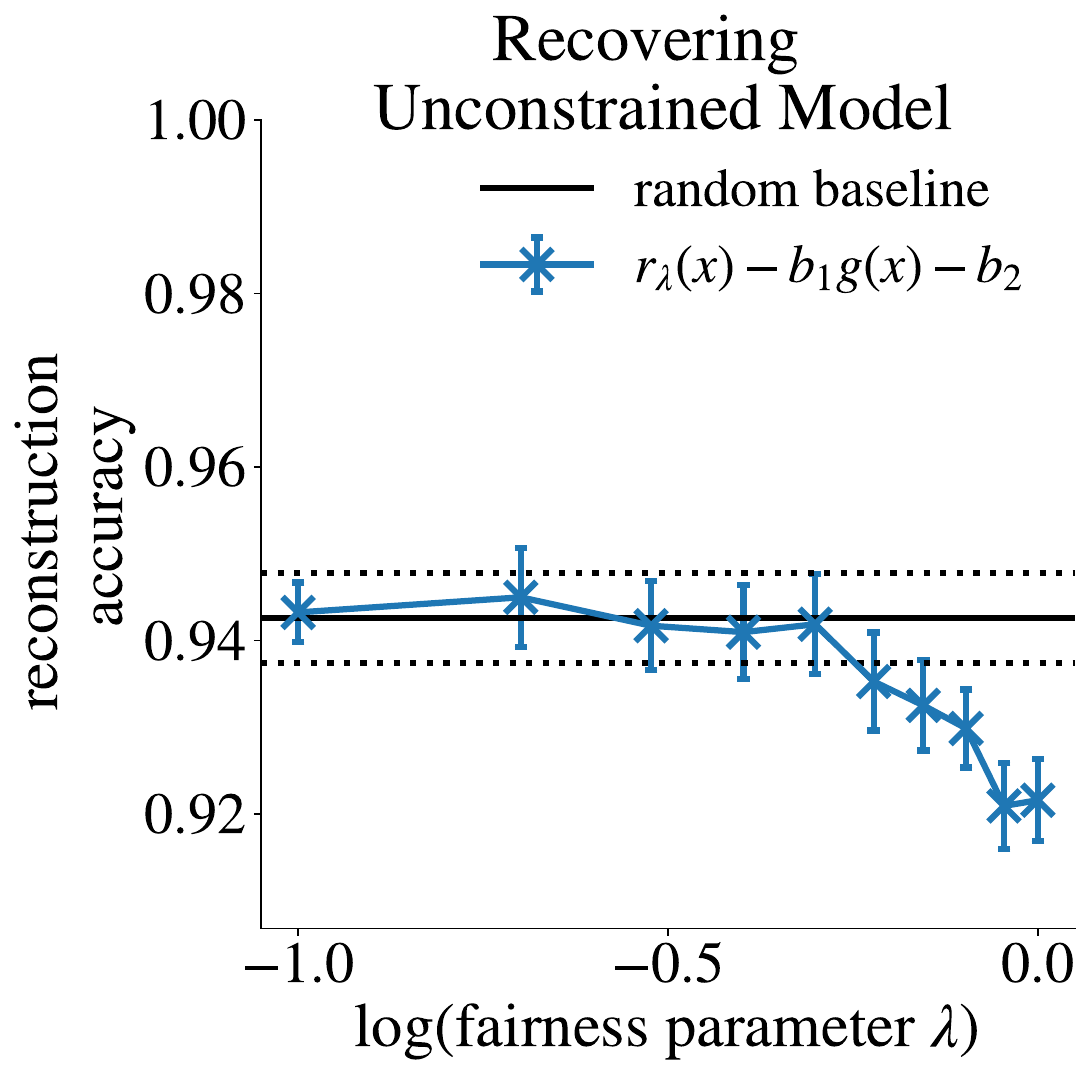}}
    \caption{\textbf{Recovering the unconstrained classifier.} For different models and fairness approaches for the target \textsc{Attractive} and protected attribute \textsc{Male}, we evaluate how our approach can reproduce the behavior of the fair model.  From regularized ResNet50 models we can recover the unconstrained model well. From fair regularized or massaged MobileNetV3-Small models we recover the unconstrained model slightly worse than a retrained unconstrained model.
    }\label{suppfig:model_equivalence_attractive}
\end{figure}

\begin{figure}
    \centering
    \subcaptionbox{ResNet50 with $\widehat{\mathcal{R}}_{\mathrm{DP}}$. }{\includegraphics[width=0.3\textwidth]{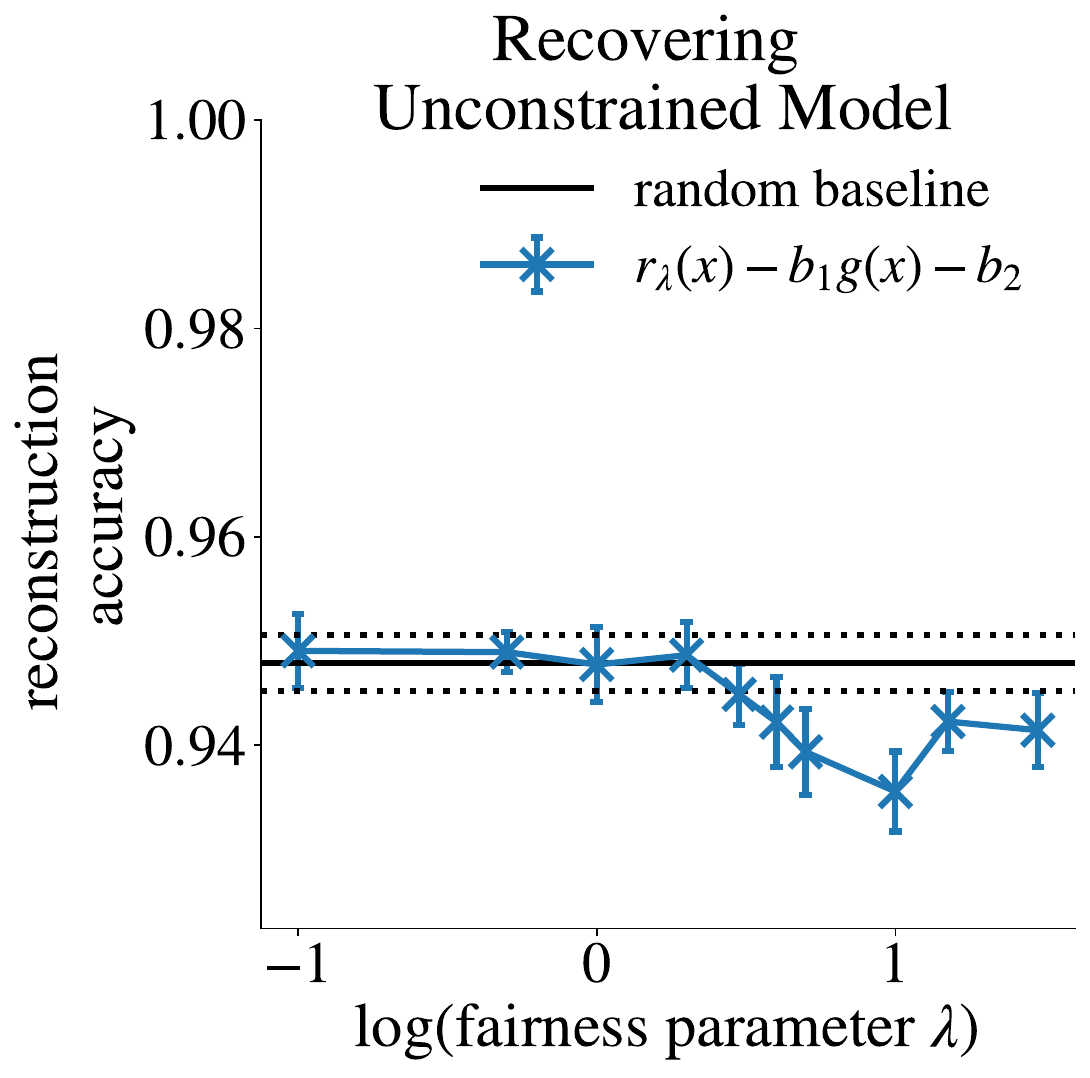}}
    \subcaptionbox{MobileNetV3-Small with $\widehat{\mathcal{R}}_{\mathrm{DP}}$.}{    \includegraphics[width=0.3\textwidth]{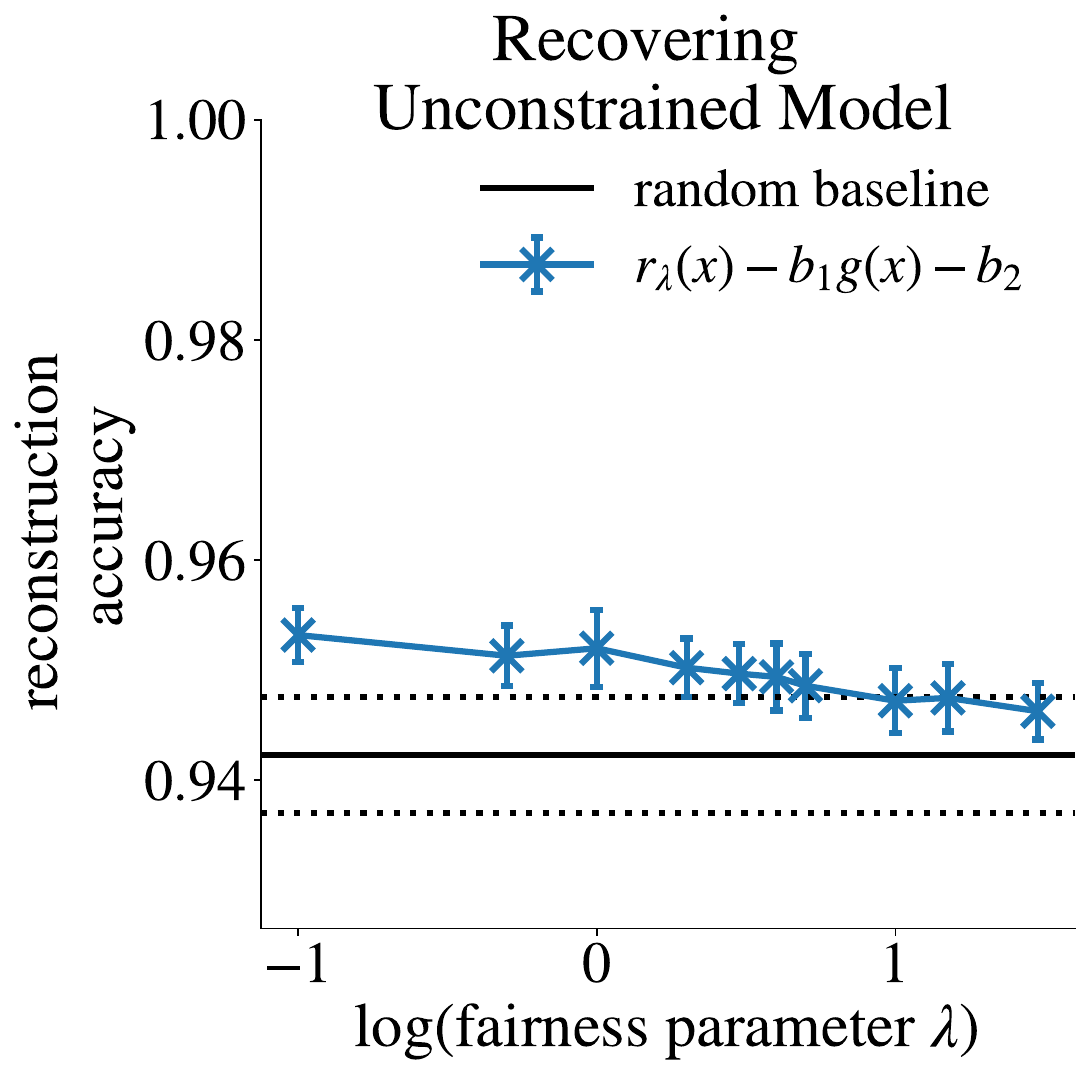}}
    \subcaptionbox{MobileNetV3-Small with \emph{Massaging}.}{
    \includegraphics[width=0.3\textwidth]{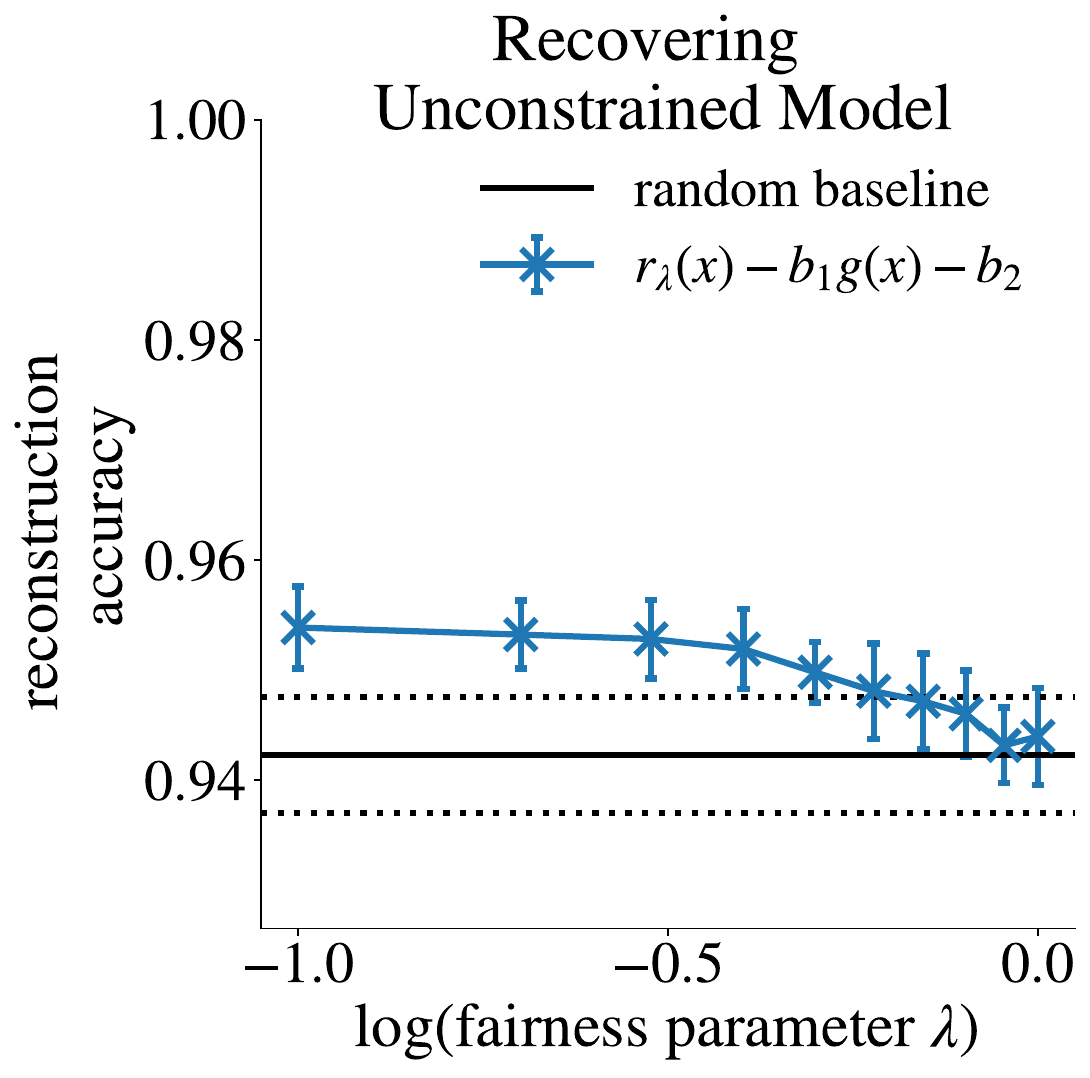}}
    \caption{\textbf{Evaluating reconstruction accuracy of our  approach.} For different models and fairness approaches for the target \textsc{Young} and protected attribute \textsc{Male}, we evaluate how close our approach is. In this example, we can recover the unconstrained model from fair models using the second head well. The accuracy for fair ResNet50 models is below the random baseline. 
    }\label{suppfig:model_equivalence_young}
\end{figure}

In this section, we conduct the experiments from Section~\ref{sec:model_equivalence} on other tasks and computer vision models. Predicting \textsc{Smiling} we use our  approach to reconstruct fair models and recover the unconstrained model using a ResNet50 with $\widehat{\mathcal{R}}_{\mathrm{DP}}$ regularizer (Figure~\ref{suppfig:model_equivalence_smiling_resnet}), using a MobileNetV3-Small with $\widehat{\mathcal{R}}_{\mathrm{DP}}$ regularizer (Figure~\ref{suppfig:model_equivalence_smiling_mobile}), and using a MobileNetV3-Small with \emph{Massaging} preprocessing (Figure~\ref{suppfig:model_equivalence_smiling_mobile_massaging}). In Figure~\ref{suppfig:model_equivalence_attractive} and \ref{suppfig:model_equivalence_young}, we recover the unconstrained model from fair ResNet50 and MobileNetV3-Small models predicting either \textsc{Attractive} or \textsc{Young}. Overall, we are able to replicate the behavior of fair models using both heads of our explicit approach and to recover the unconstrained model from a fair model with the group classifier head. Sometimes, as observed in Figure~\ref{suppfig:model_equivalence_attractive} the unconstrained classifier cannot be recovered from the fairest models within the performance of the random baseline.


\newpage

\FloatBarrier

\subsection{Disparate Treatment in Fair Networks.}\label{suppsec:disparate_treatment}

\begin{figure}[h!]
    \centering\subcaptionbox{MobileNetV3-Small with regularizer $\widehat{\mathcal{R}}_{\mathrm{DP}}$.}{
    \includegraphics[width=0.4\textwidth]{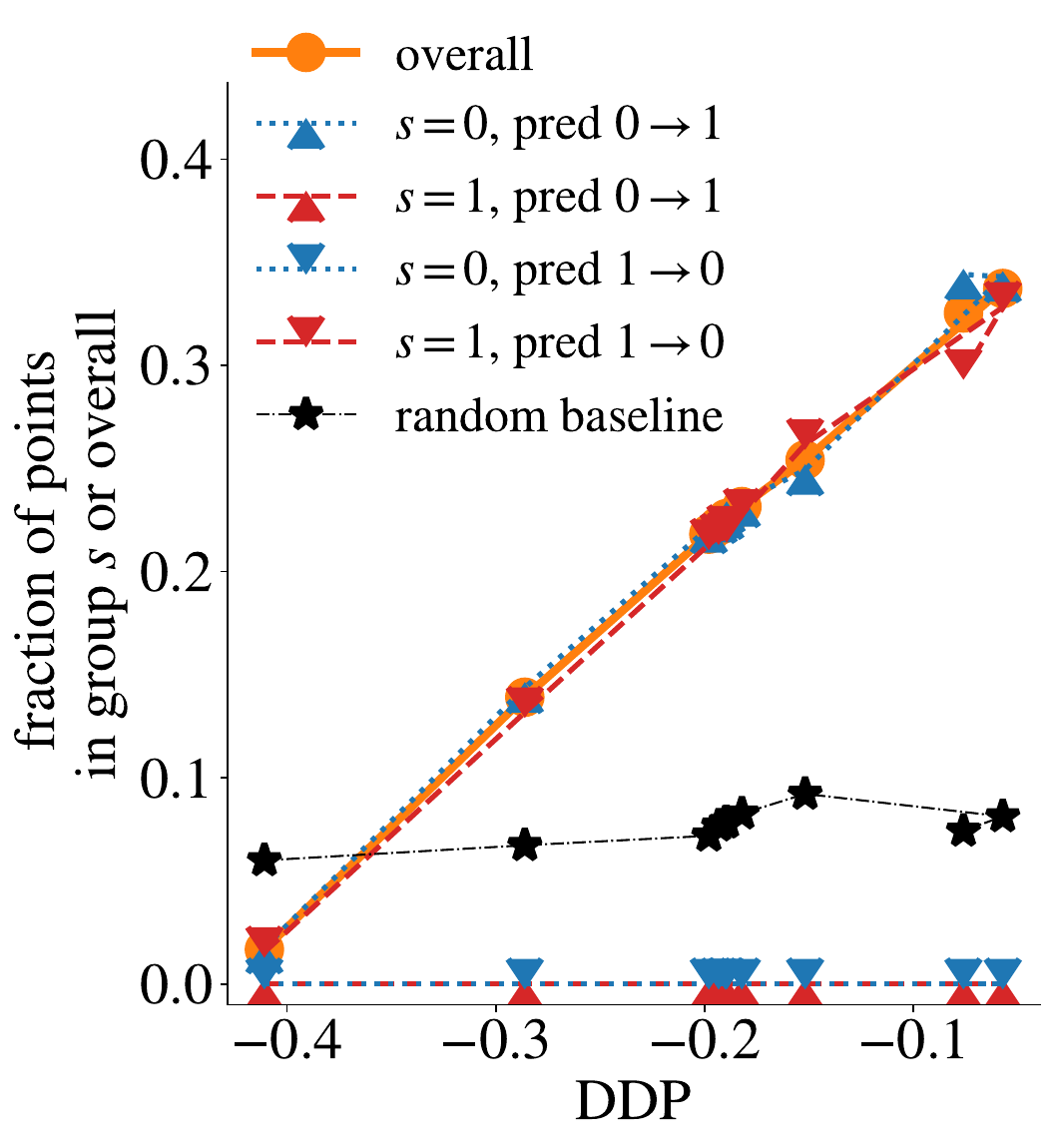}}\hspace{1cm}
    \subcaptionbox{MobileNetV3-Small with \emph{Massaging} preprocessing.}{\includegraphics[width=0.4\textwidth]{plots/disparate_treatment/Male_Attractive/mobilenetv3_small_massaging_counterfactuals_v2.pdf}}
    \caption{We perform our analysis described in Section~\ref{sec:disparate_treatment} on CelebA with target attribute \textsc{Attractive} and protected attribute \textsc{Male}. For both the regularizer and the preprocessing, up to $35\%$ of all points would receive a different outcome if their inferred attribute changed. As expected, only in one group negative predictions change into positive predictions; at the same time only for the other group positive prediction change to negative predictions.
    }\label{suppfig:disparate_treatment_mobilenet_attractive}
\end{figure}
\begin{figure}[h!]
    \centering
    \subcaptionbox{MobileNetV3-Small with regularizer $\widehat{\mathcal{R}}_{\mathrm{DP}}$.}{\includegraphics[width=0.4\textwidth]{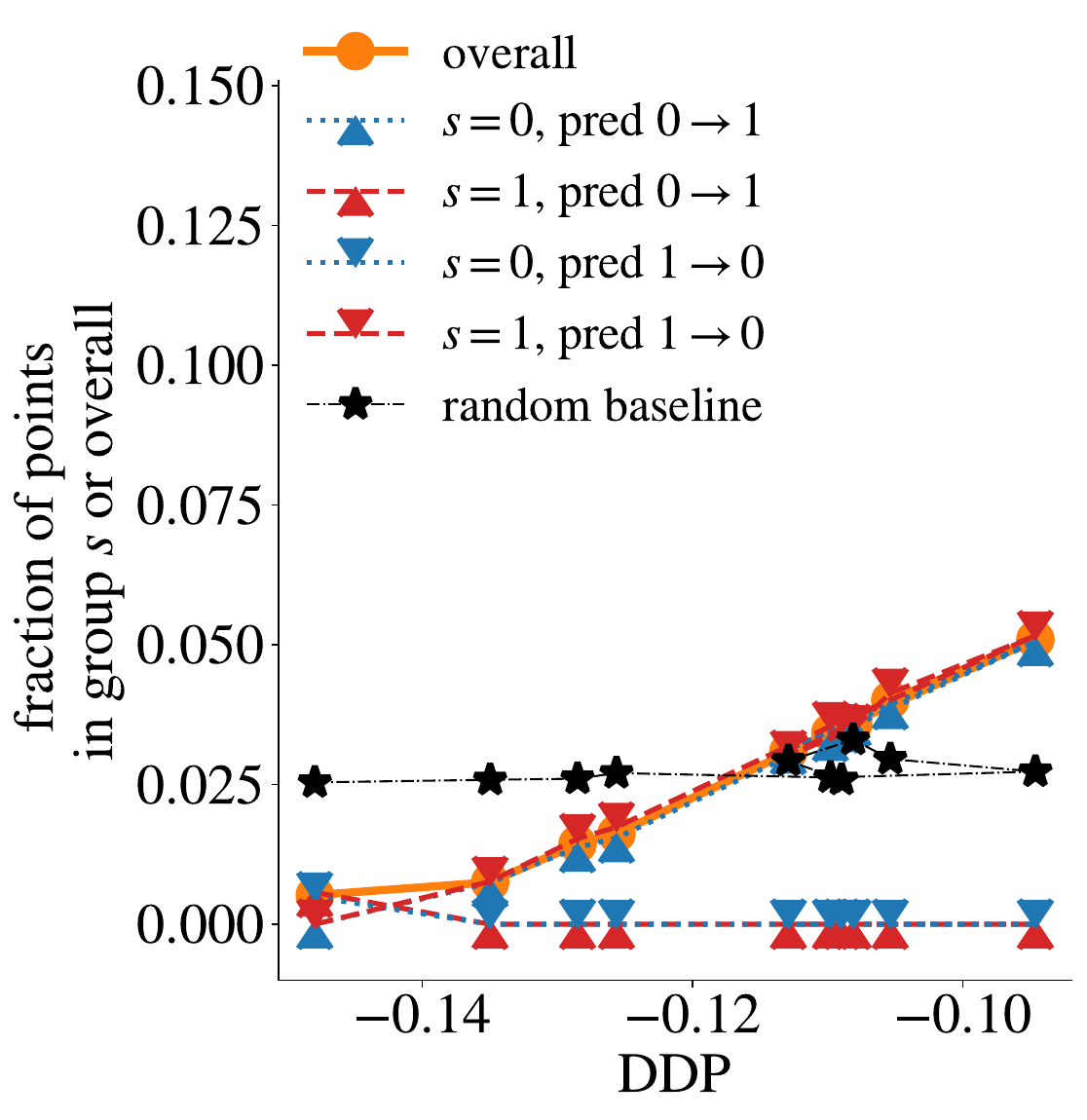}}\hspace{1cm}
    \subcaptionbox{MobileNetV3-Small with \emph{Massaging} preprocessing.}{\includegraphics[width=0.4\textwidth]{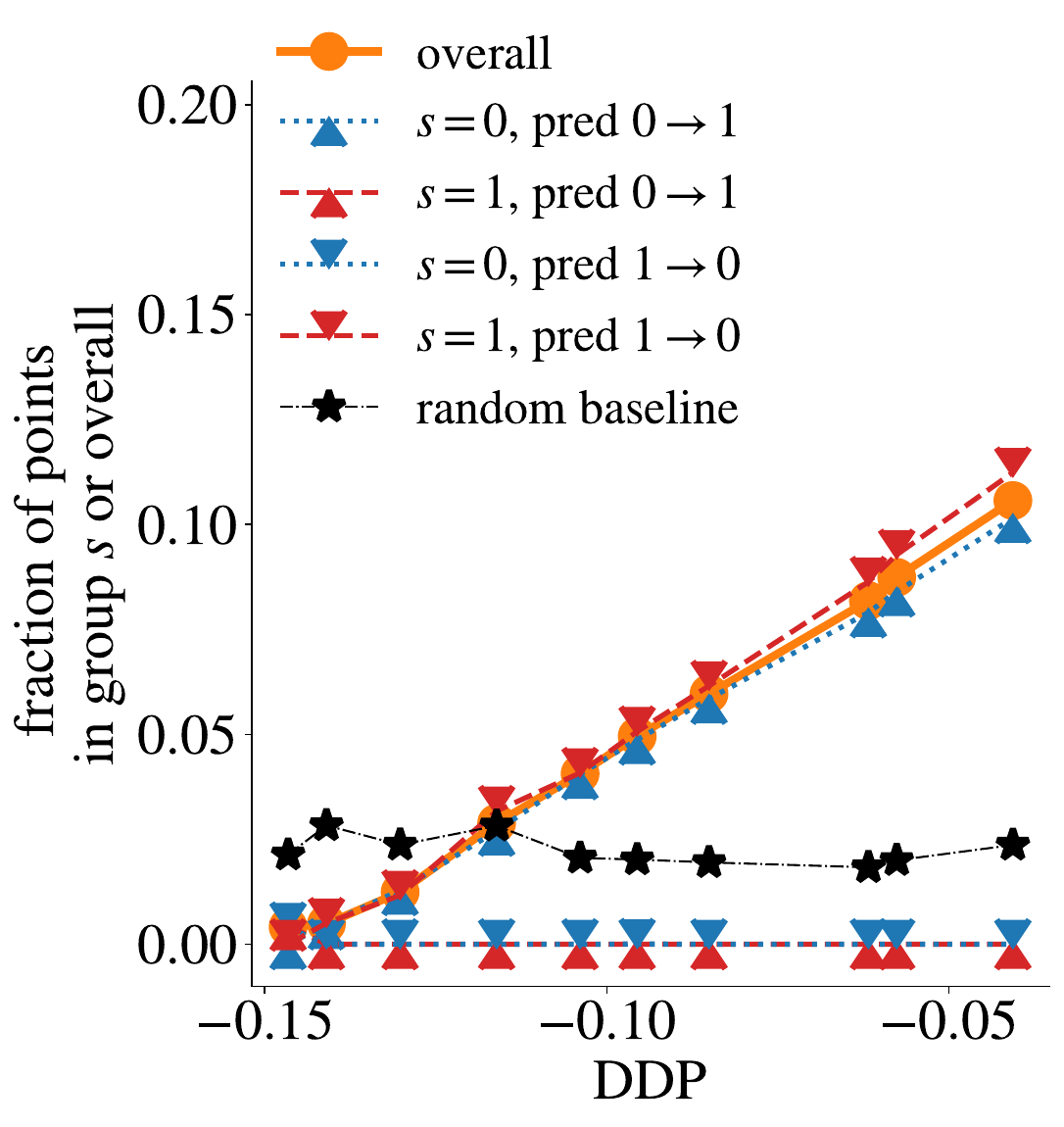}}
    \caption{We perform our analysis described in Section~\ref{sec:disparate_treatment} on CelebA with target attribute \textsc{Smiling} and protected attribute \textsc{Male}. We trained MobileNetV3-Small models with (Left) the $\widehat{\mathcal{R}}_{\mathrm{DP}}$ regularizer, or (Right) the \emph{Massaging} preprocessing method. We observe a similar behavior as above. Since the regularized approach is not improving fairness further than $\mathrm{DDP} = -0.1$, the fraction of changed predictions is only slightly higher than the baseline.
    }\label{suppfig:disparate_treatment_mobilenet_smiling}
\end{figure}

\end{document}